\pdfoutput=1
\documentclass[letterpaper]{article} 
\usepackage[]{emnlp}  
\usepackage{times}  
\usepackage{helvet}  
\usepackage{courier}  
\usepackage{graphicx} 
\usepackage{natbib}  
\usepackage{caption} 
\usepackage{algorithmic}
\usepackage[linesnumbered, ruled]{algorithm2e}

\usepackage[utf8]{inputenc} 
\usepackage[T1]{fontenc}    
\usepackage{url}            
\usepackage{booktabs}       
\usepackage{multirow}
\usepackage{amsfonts}       
\usepackage{nicefrac}       
\usepackage{microtype}      
\usepackage{xcolor}         
\usepackage{wrapfig}
\usepackage[inline]{enumitem}
\usepackage{amsmath}
\usepackage{graphicx}
\usepackage{wrapfig}
\usepackage{mathabx}
\usepackage{caption}
\usepackage{subcaption}

\title{Self-Convinced Prompting: Few-Shot Question Answering with Repeated Introspection}

\author{%
  Haodi Zhang$^1$\\
  \And
  Min Cai$^1$\\
  \And
  Xinhe Zhang$^1$\\
  \AND
  Chen Jason Zhang$^2$\\
  \And
  Rui Mao$^1$\\
  \And 
  Kaishun Wu$^3$\\
  \AND
  \textmd{Shenzhen University}$^1$ \ \
  \textmd{The Hong Kong Polytechnic University}$^2$ \\
  \textmd{The Hong Kong University of Science and Technology (Guangzhou)}$^3$
  }

\begin{document}

\maketitle

\begin{abstract}

While large language models (LLMs) such as ChatGPT and PaLM have demonstrated remarkable performance in various language understanding and generation tasks, their capabilities in complex reasoning and intricate knowledge utilization still fall short of human-level proficiency. Recent studies have established the effectiveness of prompts in steering LLMs towards generating desired outputs. Building on these insights, we introduce a novel framework that harnesses the potential of large-scale pre-trained language models, to iteratively enhance performance of the LLMs. Our framework incorporates three components: \textit{Normal CoT}, a \textit{Convincer}, and an \textit{Answerer}. It processes the output of a typical few-shot chain-of-thought prompt, assesses the correctness of the response, scrutinizes the answer, refines the reasoning, and ultimately produces a new solution. Experimental results on the 7 datasets of miscellaneous problems validate the efficacy of the Self-Convince framework, achieving substantial improvements compared to the baselines. This study contributes to the burgeoning body of research focused on integrating pre-trained language models with tailored prompts and iterative refinement processes to augment their performance in complex tasks\footnote{The code is available at: https://github.com/xxxxx.}
\end{abstract}

\section{Introduction}

Recent advancements in large-scale pre-trained language models (LLMs) have shown impressive results in various natural language understanding and generation tasks ~\cite{radford2018improving}.  However, harnessing the full potential of these models to solve complex problems, e.g., arithmetic problems, remains a challenging task. One of the cutting-edge techniques, i.e., Chain-of-Thought (CoT) prompting has achieved astonishing results by forcing LLMs to solve problems step-by-step. Another technique, named In-Context Learning (ICL), is able to guide LLMs in generating desired outputs by giving examples in the context ~\cite{gao2020making}, demonstrating the effectiveness of prompt-based approaches in various applications. Based on these findings, the latest literature endeavors to make full use of models' outputs and explore abilities beyond naive question answering. One of the abilities is to judge and analyze the reasoning paths of an answer, and amend them if necessary.

\begin{figure*}[t]
  \centering
  \includegraphics*[width=\textwidth]{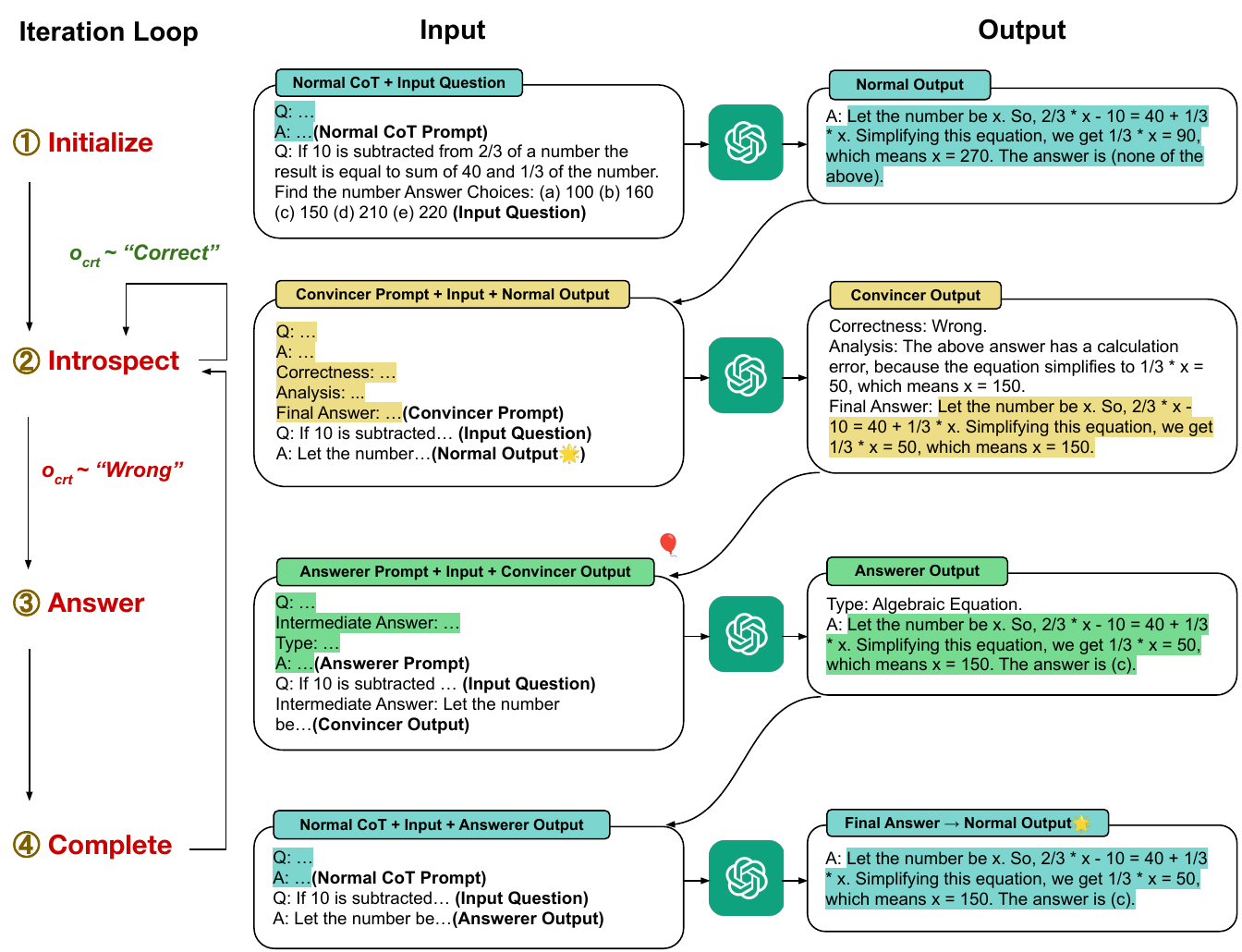}
  \caption{Framework of \textit{Self-Convince}. There are three parts shown in the figure. The left part illustrates an \textbf{``Iteration Loop''}; the middle and the right part demonstrate details of the inputs and outpus of each module. Prompts for each module are highlighted in \textbf{``Input''}, and the outputs to the next phases of each module are also highlighted in \textbf{``Ouput''}. The balloon marks where the \textit{Answerer-first Inference} takes place. The star marks the output to the next iteration.}
  \label{fig:framework}
\end{figure*}

However, prior research solely focuses on refining the reasoning paths with ground truth or heuristics as supervisory information, whereas has not explored the ability of judging the correctness, analyzing the reasoning paths, and amending the paths by LLMs themselves. We call this ability as ``self-convince'', i.e., generating self-convinced outputs based on a given input question-answer pair, alongside the reasoning paths. To this end, we introduce a novel iterative framework named \textit{Self-Convince}. As is shown in Figure~\ref{fig:framework}, it constitutes of three modules, i.e., \textit{Normal CoT}, \textit{Convincer}, and \textit{Answerer}. 

The framework operates as follows:
\begin{enumerate*}[label=(\arabic*)]
\item Given a question-answer pair generated from an arbitrary \textit{Normal CoT} (\textbf{Initialize}),
\item \label{intro:introspect} The \textit{Convincer} module first assesses the correctness of the answer and reasoning steps (\textbf{Introspect}),
\item \label{intro:answer} If the reasoning steps produce incorrect answers, the \textit{Convincer} module outputs a rectified reasoning path to the \textit{Answerer} module. The \textit{Answerer} then provides an answer based on the rectified reasoning path, along with self-hinted question type information (\textbf{Answer}).
\item \label{intro:complete} Finally, a \textit{Normal CoT} module completes the output from the last module (\textbf{Complete}).
\end{enumerate*}

Furthermore, an iteration loop is formed by feeding the output of step \ref{intro:complete} to step \ref{intro:introspect}. Experimental results in Section~\ref{sec:exp} demonstrate the effectiveness of the \textit{Self-Convince} framework, achieving state-of-the-art average improvements on multiple benchmarks and attaining the state-of-the-art performance on AQuA and SVAMP datasets.

In addition, we conduct ablation studies for each module (Section~\ref{sec:ablation}) and provide a comprehensive discussion on the observed phenomena from our experiments (Section~\ref{sec:discussion}). The statistical results on the benchmarks and the analysis of the observed phenomena aim to shed light on the ability of LLMs to autonomously "self-convince."

In summary, our contributions are fourfold:

\begin{enumerate}
    \item We propose a novel framework, named \textit{Self-Convince}, which employs a LLM to iteratively evaluate, analyze, and rectify its reasoning, and generate an answer.
    \item We propose two inference methods and an approach to construct answer choices based on our framework. We also carry out adequate ablations for our modules and methods.
    \item We conduct experiments on several benchmarks, ranging from English arithmetic, commonsense problems, to Chinese arithmetic problems. Statistical results show that our method is able to iteratively enhance the performance of an arbitrary CoT prompting, and demonstrates the potential of being applied to a broader scope of tasks.
    \item Furthermore, we carry out empirical analysis of observations from our experiments, and hope that it inspires future research to explore much more sophisticated abilities of LLMs. 
\end{enumerate}

\section{Related Work}

\subsection{In-Context Learning}
The escalation in model size and corpus dimensions ~\cite{devlin2018bert,lagler2013gpt2,brown2020language,chowdhery2022palm} has enabled Large Language Models (LLMs) to exhibit an In-Context Learning (ICL) capacity - a feature that equips Large Language Models (LLMs) with the capacity to accomplish desired tasks during inference using a handful of task-specific instances as demonstrations, all without altering the model's parameters ~\cite{shao2023synthetic,smith2022using,scao2022bloom}.
Zhao ~\shortcite{zhao2021calibrate} profoundly emphasized the critical role that example selection and arrangement play in the efficacy of LLMs in an ICL environment. Moreover, the utilization of such demonstrations fosters a deeply intertwined relationship between Chain of Thought (CoT) prompting and ICL, thereby propelling considerable research interest toward devising strategies for the identification of fitting few-shot demonstrations.
Since the advent of few-shot prompting, as introduced by ~\cite{brown2020language}, numerous methodologies have emerged to enhance the prompting capabilities of models. These include automated prompt learning ~\cite{lester2021power} and providing models with task-specific instructions ~\cite{wei2021finetuned, sanh2021multitask, ouyang2022training}. Moreover, the utilization of demonstrations has forged a profound correlation between  In-Context Learning (ICL) and Chain of Thought (CoT) prompting, which will be introduced in the next subsection.

\subsection{Chain-of-Thought Prompting}
With the advancement of large-scale language models (LLMs) and In-Context Learning (ICL), particularly a series of novel prompt-related works, natural language models have achieved new breakthroughs in many NLP downstream tasks involving reasoning and decision-making. CoT prompting is a gradient-free technique that induces LLMs to generate intermediate reasoning steps leading to the final answer. Wei et al. ~\shortcite{wei2022chain} formally investigated the topic of CoT prompting in language models. This method encourages LLMs to produce a coherent sequence of intermediate reasoning steps, culminating in the final response to a question. Research has demonstrated that LLMs can perform CoT reasoning with zero-shot prompting or manually written few-shot demonstrations ~\cite{kojima2022large,wei2022chain}. 
Although CoT and related works have shown outstanding performance in many traditional NLP reasoning and decision-making tasks, there are still limitations when faced with complex logical reasoning or multi-hop problems.

ReAct ~\cite{yao2022react} investigates the integration of LLMs for generating both reasoning traces and task-specific actions simultaneously, fostering increased synergy between them. This approach constitutes a general framework for combining reasoning and action using language models to tackle a variety of language reasoning and decision-making tasks. The Describe, Explain, Plan, and Select (DEPS) method employs multi-step reasoning and sub-task error rectification to address long-range tasks ~\cite{wang2023describe}. Although DEPS showcases notable performance by explaining errors in sub-tasks during trials, it depends on instant failure detection for subtasks and is unable to account for errors that may have emerged across an extensive range of actions and subtasks.

Additionally, the work on DERA ~\cite{nair2023dera} is intriguing and insightful. With the emergence of GPT-4 ~\cite{openai2023gpt4}, which is capable of robust and realistic conversation, it employs dialogue as the medium for interaction. This approach frames the dialogue as a discussion between two agent types – a Researcher, who processes information and identifies essential problem components, and a Decider, who possesses the autonomy to integrate the Researcher's information and make judgments on the final output.

Self-Consistency ~\cite{wang2022self} adopts a synthesis process that resembles multiple samplings from the same Chain-of-Thought prompt. It initially samples a diverse set of reasoning paths rather than solely relying on the greedy approach, subsequently selecting the most consistent answer by marginalizing the sampled reasoning paths.

\subsection{Iterative Approach to Enhance Outputs from Large Language Models}

In contrast to Self-Consistency, there is a line of work which focuses on improving outputs from LLMs using iterative approaches. Self-Refine ~\cite{madaan2023self} proposes a framework that aims to enhance the initial outputs of LLMs through iterative feedback and refinement. The key concept revolves around utilizing an LLM to generate an output, obtaining multi-aspect feedback from the same model regarding its own output, and subsequently refining the previously generated output based on this feedback. Notably, this iterative refinement framework does not require supervised training data or reinforcement learning, and it operates solely with a single LLM.

Reflexion ~\cite{shinn2023reflexion} represents a significant advancement over previous approaches such as ReAct and DEPS. By employing a binary reward model, Reflexion equips agents with dynamic memory and self-reflection capabilities, effectively augmenting their reasoning trace and task-specific action selection abilities. Meanwhile, Iter-CoT adopts an iterative strategy to enhance reasoning steps and answers by utilizing the correct answer as supervisory information, ultimately generating exemplars through sampling from the boosted examples. Similarly, PHP employs iterative techniques by leveraging hints from previous loops to generate answers until no new answers are generated.

In this paper, we conduct a comprehensive comparison of our proposed methods with Iter-CoT and PHP, both of which utilize additional supervisory information to guide and terminate the iteration process. In contrast, our method relies solely on the intrinsic capabilities of LLMs during the iteration, showcasing its distinctive approach and highlighting its potential advantages.



\section{Self-Convince Framework}

\begin{algorithm}[t]
\SetKwInOut{Input}{Input}\SetKwInOut{Output}{Output}\SetKwInOut{Require}{Require}
\Input{Question $x$}
\Output{Answer $a$} 
\Require{Normal-CoT $f_P$, Convincer $f_C$, \\Answerer $f_A$}
\begin{algorithmic}[1]
\STATE $r_{0}, a_{0} \gets f_{P}(x)$\;
\FOR{$i \in$ 1 to $n$}
\STATE $o_{crt},o_{aly},r_c \gets f_{C}(r_{i-1}, a_{i-1})$
\STATE $o_{typ},r_a \gets f_{A}(r_c)$
\IF{$o_{crt} \sim $\textit{``Correct''}}
\STATE \textit{continue}
\ENDIF
\IF{\textit{Normal Inference}}
\STATE $\hat{x} \gets x \oplus r_a$ \label{alg:normal_inference}
\ELSIF{\textit{Answerer-first Inference}}
\STATE $\hat{x} \gets x \oplus o_{typ}$ \label{alg:answerer_inference}
\ENDIF
\STATE $r_{i}, a_{i} \gets f_{P}(\hat{x})$
\ENDFOR
\RETURN $a_{n}$
\caption{Algorithm of Self-Convince}
\label{alg:iteration}
\end{algorithmic}
\end{algorithm}

We hereby present our novel framework, titled \textit{Self-Convince}, which is illustrated in Figure~\ref{fig:framework}. The framework comprises three fundamental components, namely \textit{Normal CoT''}, \textit{Convincer''}, and \textit{Answerer''}. These modules collectively facilitate four sequential steps, namely ``Initialize'', ``Introspect'', ``Answer'', and ``Complete''.

During the initial phase, the \textit{Normal CoT} generates an initial answer, referred to as the Normal Output,'' for the input question. This output is subsequently provided as input to the \textit{Convincer}. Detailed insights into the functioning of the \textit{Convincer} can be found in Figure~\ref{fig:framework}, where it yields three key outputs: Correctness,'' Analysis,'' and Final Answer,'' based on the input question and the Normal Output.'' If the model deems the answer correct, as indicated by \textit{Correctness: Correct''} in the Convincer Output,'' the original answer from the ``Normal Output'' is retained as the final answer, and the ``Introspect'' step is reiterated.

Conversely, if the \textit{Convincer} determines the answer as incorrect, the ``Final Answer'' from the ``Convincer Output'' is fed into the final step, i.e., ``Answer''. This step utilizes the \textit{Answerer} to generate an intermediate answer. Finally, a \textit{Normal CoT} will be used to ``Complete'' the answer. Subsequently, the resulting answer undergoes the ``Introspect'' step as the new iteration's ``Normal Output.''

Henceforth, we have provided a high-level overview of our framework. Detailed and formal explanations will be presented in the subsequent sections.
\subsection{Normal Chain-of-Thought Prompting}
The \textit{Self-Convince} framework is initialized by utilizing an arbitrary chain-of-thought prompt denoted as $P$. By providing an input $x$ and utilizing the model $f$, the system generates an answer along with its corresponding reasoning path. This process can be formally represented as the function $f \colon (x, P)\to (r, a)$, where $r$ represents the sequence of tokens that constitutes the reasoning path, and $a$ denotes the sequence of tokens containing the answer, typically located at the conclusion of the generated output. To simplify the notation, we can represent the model with a prompt $P$ as $f_P$, resulting in the revised function notation: $f_P \colon x \to (r,a)$. Throughout the subsequent sections, we will employ similar notations for clarity and consistency.

\subsection{Marginalized Convincer}
The second crucial component of the \textit{Self-Convince} framework is the \textit{Convincer}, denoted as $f_C$, which plays a central role in the system. Taking $x$ and $(r,a)$ as input, the \textit{Convincer} module produces a modified reasoning path, where the last reasoning step is corrected. Its function can be defined as $f_C \colon (x, r, a)\to (o_{crt},o_{aly},r_c)$. The \textit{Convincer} possesses the ability to identify errors within the original reasoning path $r$, conduct an analysis of the mistakes, and subsequently rectify them. Once the erroneous reasoning step has been revised, the subsequent steps are removed from the reasoning path.

As illustrated in Figure~\ref{fig:framework}, the \textit{Convincer} module can detect an error in the reasoning path, specifically in the segment \textit{"...So, 2/3 * x - 10 = 40 + 1/3 * x. Simplifying this equation, we get 1/3 * x = 90..."} Upon analyzing the mistake, it corrects the error and outputs a truncated answer, such as \textit{"Let the number be x. So, 2/3 * x - 10 = 40 + 1/3 * x. Simplifying this equation, we get 1/3 * x = 50, which means x=150."} It is important to note that a reasoning path may contain multiple errors. Therefore, in the design of the \textit{Convincer} module, we focus on analyzing and rectifying the first identified mistake while marginalizing the remaining steps. We hope that language models can acquire this ability through learning, and we will further explore this aspect in our subsequent discussions.

\subsection{Step-wise Answerer}
The third building block of our framework is referred to as the \textit{``Answerer''}, denoted as $f_A$. It can be perceived as a chain-of-thought prompt that is compelled to engage in single-step reasoning. Taking the reasoning path derived from the \textit{Convincer} module, denoted as $\hat{r}$, as its input, the \textit{Answerer} produces supplementary information required to address the given question. Furthermore, it provides an extended reasoning path that encompasses an additional reasoning step. This can be expressed formally as the function $f_A\colon (x, \hat{r})\to (o_{typ}, r_a)$.

As demonstrated by our running example illustrated in Figure~\ref{fig:framework}, the \textit{Answerer} module appends pertinent information, such as ``\textit{Type: Algebraic Equation}'', following the answer derived from the previous step. By adopting this step-wise approach, the \textit{Answerer} incrementally contributes to the reasoning process, resulting in a comprehensive and well-supported response to the provided question.

\subsection{Iteration}

\begin{figure}[t]

        \centering
        \includegraphics[width=\linewidth]{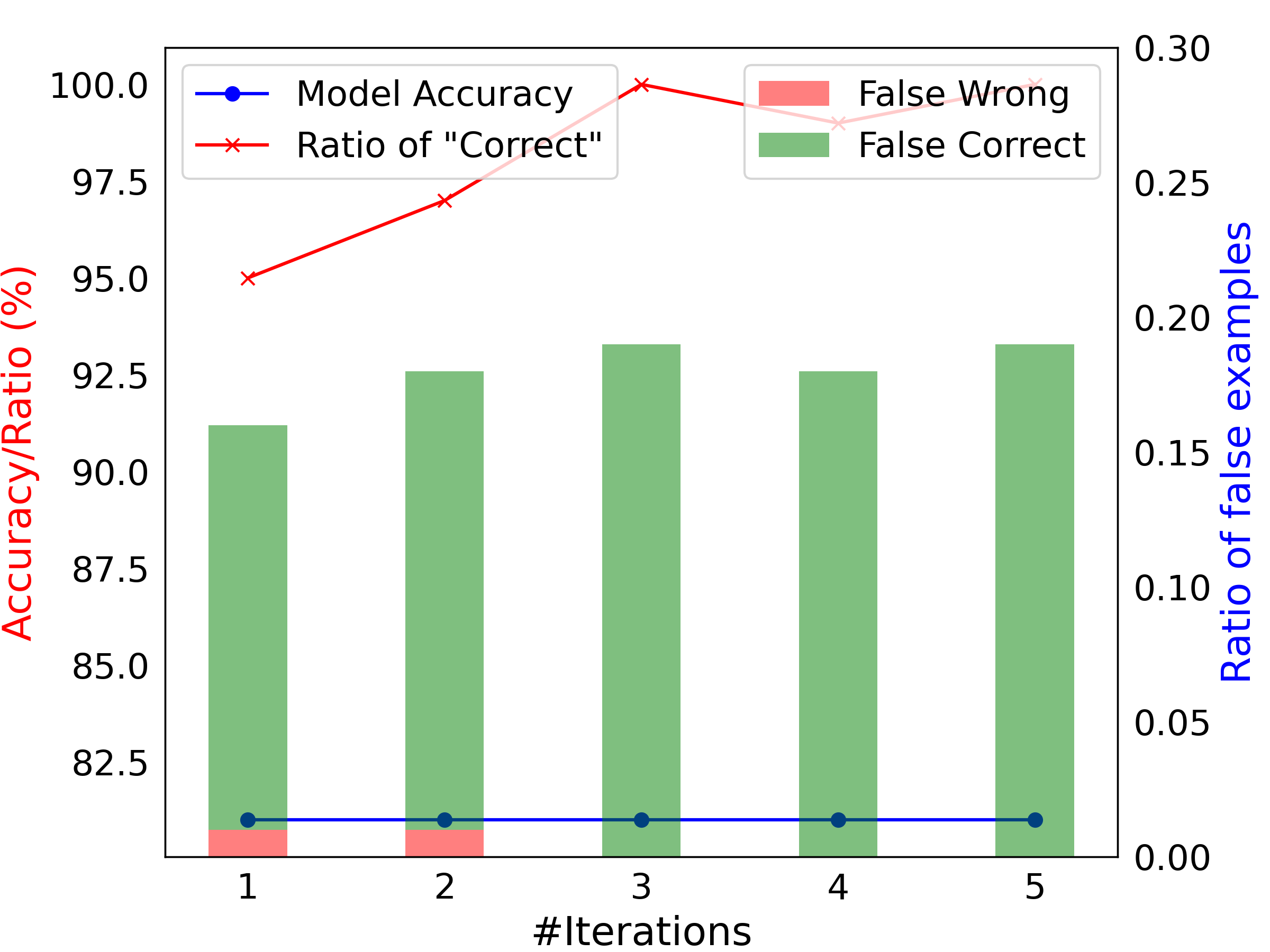}
        \caption{Convincer performance on GSM8K under normal settings.}
        \label{fig:pilot_convincer}

\end{figure}
\begin{figure}[th]

        \centering
        \includegraphics[width=\linewidth]{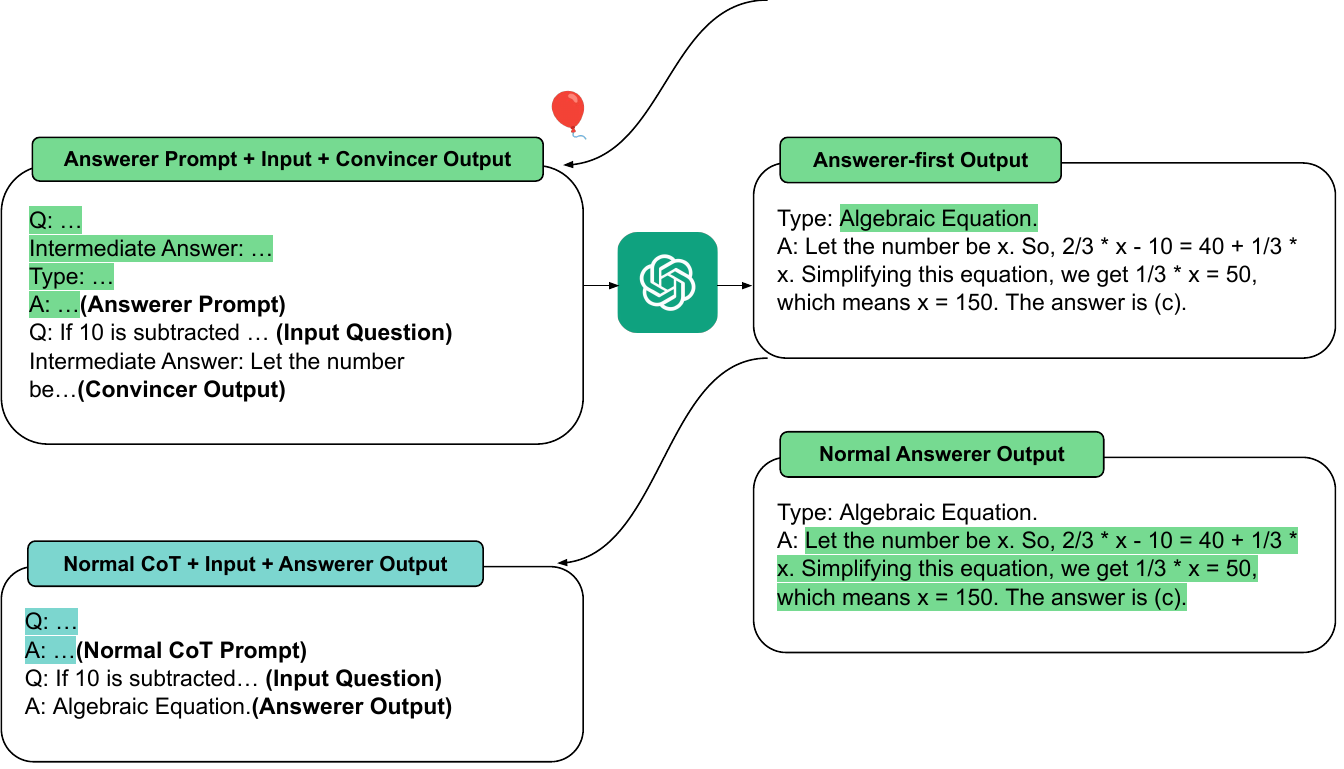}
        \caption{Answerer-first Inference.}
        \label{fig:answerer-first_inference}
\end{figure}

Upon obtaining a completed reasoning path $r_a$ from the \textit{Answerer} module, we can utilize it as the input for the \textit{Convincer} module, initiating a new iteration within the framework. This iterative process can be encapsulated by the composition of the individual functions, namely $f_{iter}\colon = f_A \circ f_C \circ f_P$. By leveraging these building blocks, the algorithm executes multiple loops, as depicted in Algorithm~\ref{alg:iteration}.

\subsection{Answer Choice Construction}
Our preliminary investigations involving AQuA, GSM8K, and SVAMP have yielded promising results, indicating the efficacy of our framework in scenarios where answer choices are provided. Additionally, we have observed that in question answering tasks without answer choices, the \textit{Convincer} module tends to produce the response \textbf{\textit{``Correctness: Correct.''}} (refer to Figure~\ref{fig:pilot_convincer}). To address this limitation, we leverage the \textit{Convincer} module to generate multiple answer choices, effectively transforming open-ended questions into closed-ended ones. This is achieved by simply appending \textit{``Wrong.''} after \textit{``Correctness:''}, thereby encouraging the model to generate a diverse range of plausible answers. We can view this as creating a ``hypothetical world'' where the correctnesses are given and serve as premises which may be false in some circumstances. For simplicity, the module is named as \textit{\textbf{Wrong-only Convincer}}.

\subsection{Answerer-first Inference}
As mentioned earlier, the \textit{Convincer} module not only provides the correctness evaluation $o_{crt}$ of the initially generated answer but also offers a corrected reasoning path $r_c$. However, expecting LLMs to modify their reasoning path can present challenges. In certain cases, the \textit{Convincer} module may accurately assess the correctness but struggle to amend the reasoning path accordingly. To address this issue, we propose an alternative inference method called \textit{Answerer-first Inference}, which avoids reliance on the corrected reasoning path $r_c$. This is achieved by incorporating the output $o_{typ}$ of the \textit{Answerer} module into the \textit{Normal-CoT} $f_P$, specifically by appending $o_{typ}$ after \textit{``A:''}, as is shown in Figure~\ref{fig:answerer-first_inference}. By adopting this approach,
 we enable the model to prioritize the \textit{Answerer} module during the inference process.

\begin{table*}[th]
	\centering
	\resizebox{\linewidth}{!}{
	\begin{tabular}{cccccccccc}
		\toprule
		\multirow{2}*{\textbf{Methods}} &
		\multicolumn{5}{c}{\textit{Arithmetic}} & \multicolumn{3}{c}{\textit{Commonsense}}\\
		& GSM8K & AddSub & SVAMP & AQuA & avg. & CSQA & Date & avg. \\
		\midrule
            \textbf{Previous Fine-tuned SOTA} & $\textrm{55.0}^\alpha$ & $\textrm{77.7}^{\beta}$ & $\textrm{57.4}^{\gamma}$ & $\textrm{37.9}^{\delta}$ & - & $\textrm{91.2}^{\sigma}$ & - & - \\
            \midrule
		\multicolumn{9}{c}{\textit{GPT-turbo-3.5-0301}} \\
            \midrule
            \textbf{Normal CoT} ~\cite{zheng2023progressive} & 82.8 & 85.5 & 81.0 & 57.4 & 76.7 & - & - & - \\
            \textbf{PHP} ~\cite{zheng2023progressive} & 85.1 & 85.3 & 83.1 & 60.6 & 78.5(+1.8) & - & - & - \\
            \midrule
            \textbf{Normal CoT} ~\cite{sun2023enhancing} & 69.3 & 86.5 & 77.2 & 47.2 & 70.1 & 77.1 & 78.6 & 77.9 \\
            \textbf{Iter-CoT(W)} ~\cite{sun2023enhancing} & 73.6 & 89.1 & 80.7 & 49.2 & 73.2(+3.1) & 76.8 & 80.0 & 78.4(+0.5) \\
            \textbf{Iter-CoT(S)} ~\cite{sun2023enhancing} & 72.9 & 85.3 & 80.1 & 52.0 & 72.6(+2.5) & 78.0 & 74.7 & 76.4(-1.5) \\
            \midrule
		\multicolumn{9}{c}{\textit{\textbf{Self-Convince} using GPT-turbo-3.5-0301}} \\
            \midrule
		\textbf{Normal CoT} & 77.3 & 78.5 & 80.9 & 55.9 & 73.2 & 75.6 & 66.7 & 71.2 \\
            \textbf{CoT w/ Wrong-only Convincer} & 80.5 & 79.3 & 84.2 & - & - & - & - & - \\
		\textbf{Normal Inference}
		 & 81.4 & 79.5 & 84.7 & 62.0 & 76.9(+3.7) & 75.6 & 66.7 & 71.2(+0.0) \\
		\textbf{Answerer-first Inference}
		 & 81.5 & 79.3 & 84.9 & 62.0 & 76.9(+3.7) & 76.5 & 66.3 & 71.4(+0.2) \\
		 \bottomrule
	\end{tabular}
	}
\caption{Main results on different datasets. Results with best improvement over normal CoT are shown in bold. Previous SOTA using fine-tuned methods shown in the table include: $\alpha$: ~\cite{cobbe2021training}; $\beta$: ~\cite{hosseini2014learning}; $\gamma$: ~\cite{pi2022reasoning}; $\delta$: ~\cite{amini2019mathqa}; $\sigma$: ~\cite{Xu_Zhu_Wang_Sun_Cheng_Liu_Gao_He_Zeng_Huang_2022}.}
\label{table:main_result}
\end{table*}
\section{Experiment}
\label{sec:exp}
In this section, we display our main results, compare the results with other methods, and carry out several ablations. 

\begin{table}[th]
\resizebox{\linewidth}{!}{
\begin{tabular}{@{}lcccccc@{}}
\toprule
Iterations       & 0    & 1    & 2    & 3    & 4    & 5    \\
\midrule
Normal Inference & 43.1 & 43.3 & 43.5 & 43.3 & 43.3 & 43.5 \\
Answerer-first   & 43.1 & 44.0 & 44.2 & 44.7 & 44.9 & \textbf{45.1} \\
\bottomrule
\end{tabular}
}
\caption{Results on arithmetic problems in the GAOKAO benchmark. Results of Iteration 0 are from  \citeauthor{zhang2023evaluating}.}
\end{table}

\begin{table*}[th]
	\centering
	\resizebox{\linewidth}{!}{
	\begin{tabular}{cccccccc}
		\toprule
		\multirow{2}*{\textbf{Methods}} &
		\multirow{2}*{\textit{Iterations}} &
		\multicolumn{4}{c}{\textit{Arithmetic}} & \multicolumn{2}{c}{\textit{Commonsense}}\\
		&& GSM8K & AddSub & SVAMP & AQuA & CSQA & Date \\
            \midrule
		\textbf{Normal CoT} & 0 & $\textrm{77.3}_{\pm0.5}$ & $\textrm{78.5}_{\pm0.7}$ & $\textrm{80.9}_{\pm1.6}$ & $\textrm{55.9}_{\pm1.1}$ & $\textrm{75.6}_{\pm0.8}$ & $\textrm{66.7}_{\pm3.0}$ \\
            \textbf{CoT w/ Wrong-only Convincer} & 0 & $\textrm{80.5}_{\pm0.7}$ & $\textrm{79.3}_{\pm1.0}$ & $\textrm{84.2}_{\pm0.8}$ & - & - & - \\
		\multirow{5}{3cm}{\centering\textbf{Self-Convince}\\(Normal Inference)}
		& 1 & $\textrm{81.0}_{\pm0.8}$ & $\textrm{79.3}_{\pm1.1}$ & $\textrm{84.5}_{\pm1.1}$ & $\textrm{59.3}_{\pm1.3}$ & $\textrm{75.5}_{\pm0.1}$ & $\textrm{66.8}_{\pm2.5}$ \\ 
		& 2 & $\textrm{81.1}_{\pm0.9}$ & $\textrm{79.3}_{\pm1.1}$ & $\textrm{84.5}_{\pm1.1}$ & $\textrm{61.0}_{\pm2.3}$ & $\textrm{75.5}_{\pm0.3}$ & $\textrm{66.7}_{\pm1.9}$ \\
		& 3 & $\textrm{81.0}_{\pm1.1}$ & $\textrm{79.4}_{\pm1.3}$ & $\textrm{84.6}_{\pm1.1}$ & $\textrm{61.8}_{\pm1.1}$ & $\textrm{75.6}_{\pm0.4}$ & $\textrm{66.6}_{\pm1.8}$ \\
		& 4 & $\textrm{81.3}_{\pm1.1}$ & $\textrm{79.5}_{\pm1.4}$ & $\textrm{84.6}_{\pm1.1}$ & $\textrm{61.8}_{\pm1.1}$ & $\textrm{75.6}_{\pm0.4}$ & $\textrm{66.7}_{\pm1.9}$ \\
		& 5 & $\textrm{81.4}_{\pm1.0}$ & $\textrm{79.5}_{\pm1.4}$ & $\textrm{84.7}_{\pm1.1}$ & $\textrm{62.0}_{\pm2.0}$ & $\textrm{75.6}_{\pm0.4}$ & $\textrm{66.7}_{\pm1.9}$ \\
            \midrule
		\multirow{5}{3cm}{\centering\textbf{Self-Convince}\\(Answerer-first)}
		& 1 & $\textrm{81.4}_{\pm0.9}$ & $\textrm{79.3}_{\pm1.1}$ & $\textrm{84.8}_{\pm0.8}$ & $\textrm{61.6}_{\pm2.0}$ & $\textrm{76.0}_{\pm0.6}$ & $\textrm{66.3}_{\pm2.1}$ \\ 
		& 2 & $\textrm{81.5}_{\pm0.8}$ & $\textrm{79.3}_{\pm1.1}$ & $\textrm{84.7}_{\pm0.9}$ & $\textrm{62.2}_{\pm1.7}$ & $\textrm{76.3}_{\pm0.6}$ & $\textrm{66.3}_{\pm2.1}$ \\
		& 3 & $\textrm{81.6}_{\pm1.0}$ & $\textrm{79.3}_{\pm1.1}$ & $\textrm{85.0}_{\pm1.2}$ & $\textrm{62.2}_{\pm1.1}$ & $\textrm{76.3}_{\pm0.8}$ & $\textrm{66.3}_{\pm2.1}$ \\
		& 4 & $\textrm{81.5}_{\pm0.9}$ & $\textrm{79.3}_{\pm1.1}$ & $\textrm{84.8}_{\pm1.3}$ & $\textrm{62.2}_{\pm0.6}$ & $\textrm{76.4}_{\pm0.8}$ & $\textrm{66.4}_{\pm2.0}$ \\
		& 5 & $\textrm{81.5}_{\pm1.1}$ & $\textrm{79.3}_{\pm1.1}$ & $\textrm{84.9}_{\pm1.2}$ & $\textrm{62.0}_{\pm0.8}$ & $\textrm{76.5}_{\pm0.7}$ & $\textrm{66.3}_{\pm2.1}$ \\
		 \bottomrule
	\end{tabular}
	}
\caption{Averaged accuracies from each steps along with the their standard deviations. We run twice to calculate the results.}
\label{table:step_results}
\end{table*}

\begin{table}[th]
\centering
\begin{tabular}{@{}cccc@{}}
\toprule
Iterations & GSM8K & AddSub & SVAMP \\ \midrule
0          & $\textrm{77.0}_{\pm0.4}$      & $\textrm{78.5}_{\pm0.7}$       &  $\textrm{80.9}_{\pm1.6}$     \\
1          & $\textrm{79.2}_{\pm0.0}$      & $\textrm{79.6}_{\pm0.1}$       &  $\textrm{83.3}_{\pm0.4}$     \\
2          & $\textrm{80.4}_{\pm0.2}$      & $\textrm{79.2}_{\pm0.0}$       &  $\textrm{84.4}_{\pm0.6}$     \\
3          & $\textrm{80.5}_{\pm0.7}$      & $\textrm{79.3}_{\pm1.0}$       &   $\textrm{84.2}_{\pm0.8}$    \\ \bottomrule
\end{tabular}
\caption{Results on Wrong-only Convincer with different iterations}
\label{table:wrong_only_steps}
\end{table}

\subsection{Experimental Setup}
In our evaluation, we apply the \textit{Self-Convince} framework to a diverse range of benchmarks, encompassing the following datasets:
\begin{enumerate*}[label=(\arabic*)]
 \item \textbf{GSM8K} \cite{cobbe2021training}, a benchmark focused on math word problems,
 \item \textbf{AddSub} \cite{hosseini2014learning}, a collection of addition and subtraction problems,
 \item \textbf{SVAMP} \cite{patel2021nlp}, a dataset comprising challenging math word problems,
 \item \textbf{AQuA} \cite{ling2017program}, a dataset consisting of algebraic word problems,
 \item \textbf{CSQA} \cite{talmor2018commonsenseqa}, a dataset featuring commonsense problems,
 \item \textbf{Date} Understanding \cite{srivastava2022beyond}, a commonsense dataset focused on date inference problems,
 \item and \textbf{GAOKAO} \cite{zhang2023evaluating}, a benchmark based on the \textit{Chinese University Entrance Examination}, encompassing various problem types.
\end{enumerate*}
For our experiments, we utilize temperature sampling with a temperature value of $T = 0.7$ consistently across all results.

We conduct our tests on the development set of CSQA, while limiting the evaluation to math problems within the GAOKAO benchmark. In particular, the specific prompts we employ are provided in the appendix. For the GAOKAO benchmark, we adopt a one-shot \textit{Convincer} approach combined with a zero-shot \textit{Answerer} method. All the ablations will be carryied on AQuA, unless explicit stated otherwise.

\subsection{Main Results}

The main results of \textit{Self-Convince} and comparison with other methods are shown in Table~\ref{table:main_result}. We report results from the last iteration (5 iterations at maximum), and scores for each iterations are shown in Table~\ref{table:step_results} and Table~\ref{table:wrong_only_steps} Results from each iteration step of the two inference types are displayed. Our methods start with the results of Manual CoT, which is marked as the \textit{0th iteration}. We consider several prior works for comparison. Two methods using ``GPT-turbo-3.5'' are included, along with their results using ``Manual CoT''. Because there is a great gap between initial results of the listed methods, we also put the average of the results and the improvement on the table. It is worth mentioning that, even though our initial result on AQuA is lower than PHP, \textit{Self-Convince} still surpass their final results by a large margin.

\begin{figure*}[t]
\begin{subfigure}[t]{0.33\linewidth}
	\centering
	\includegraphics[width=\linewidth]{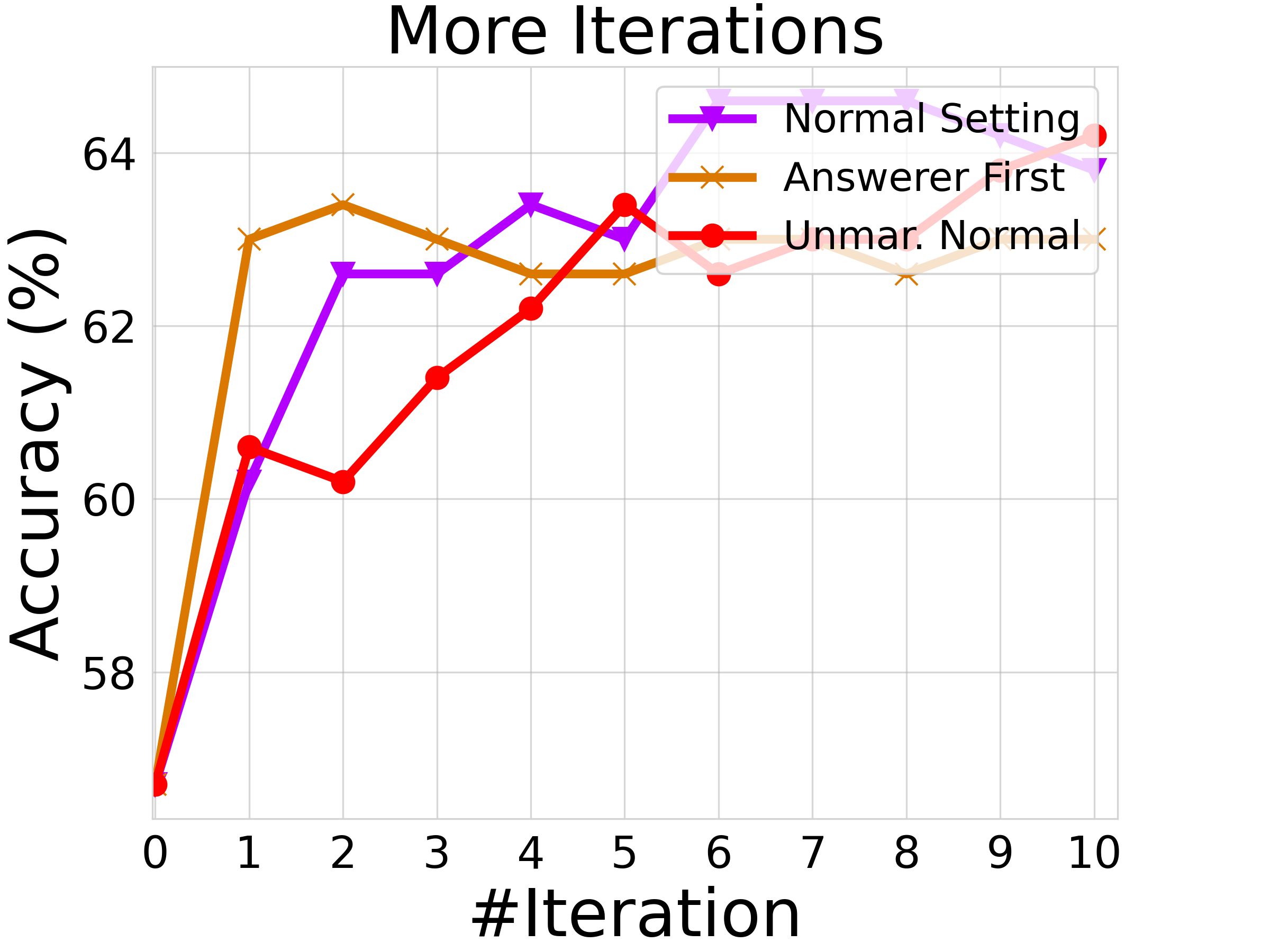}
\end{subfigure}
\begin{subfigure}[t]{0.33\linewidth}
	\centering
	\includegraphics[width=\linewidth]{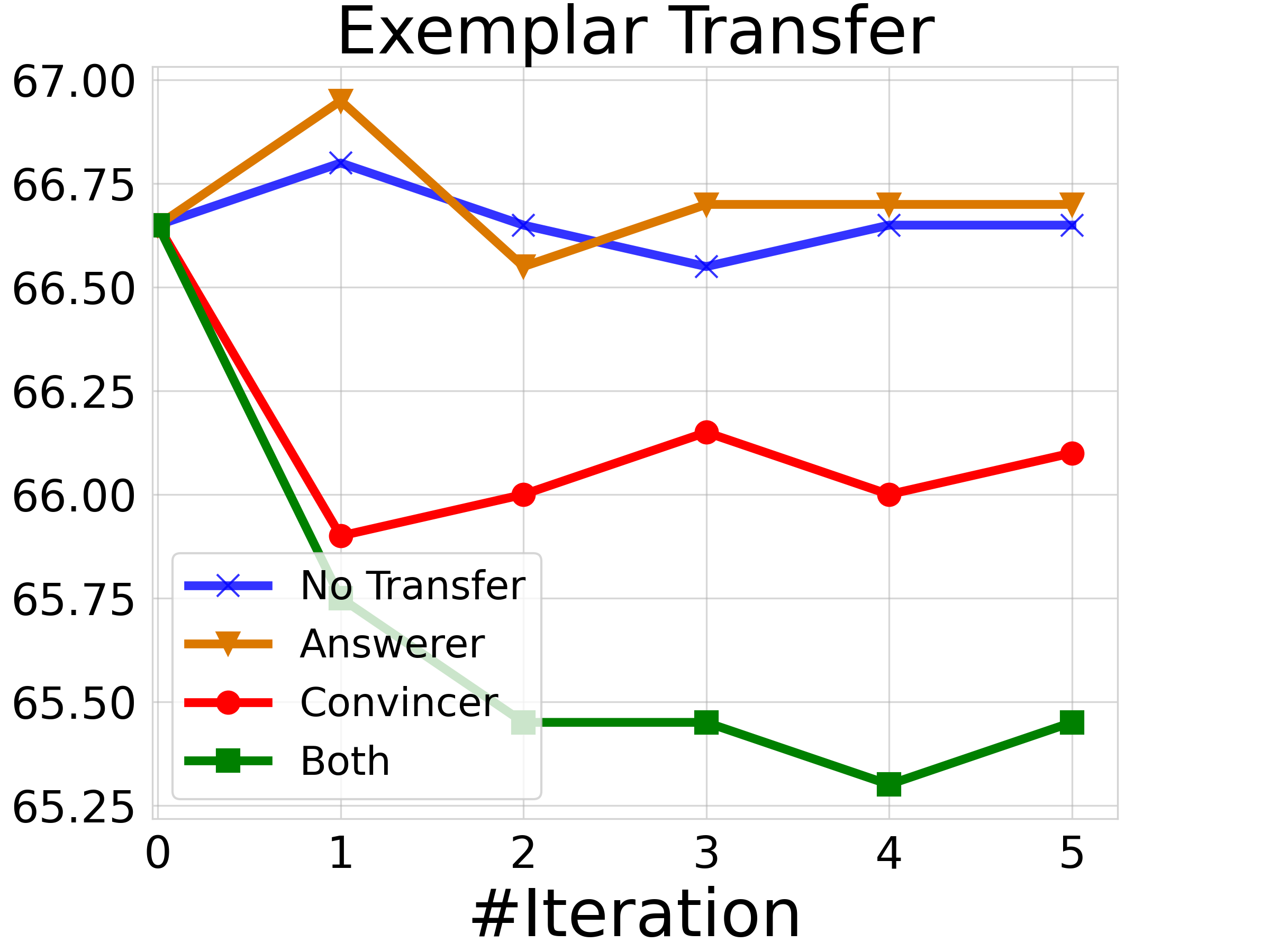}
\end{subfigure}
\begin{subfigure}[t]{0.33\linewidth}
	\centering
	\includegraphics[width=\linewidth]{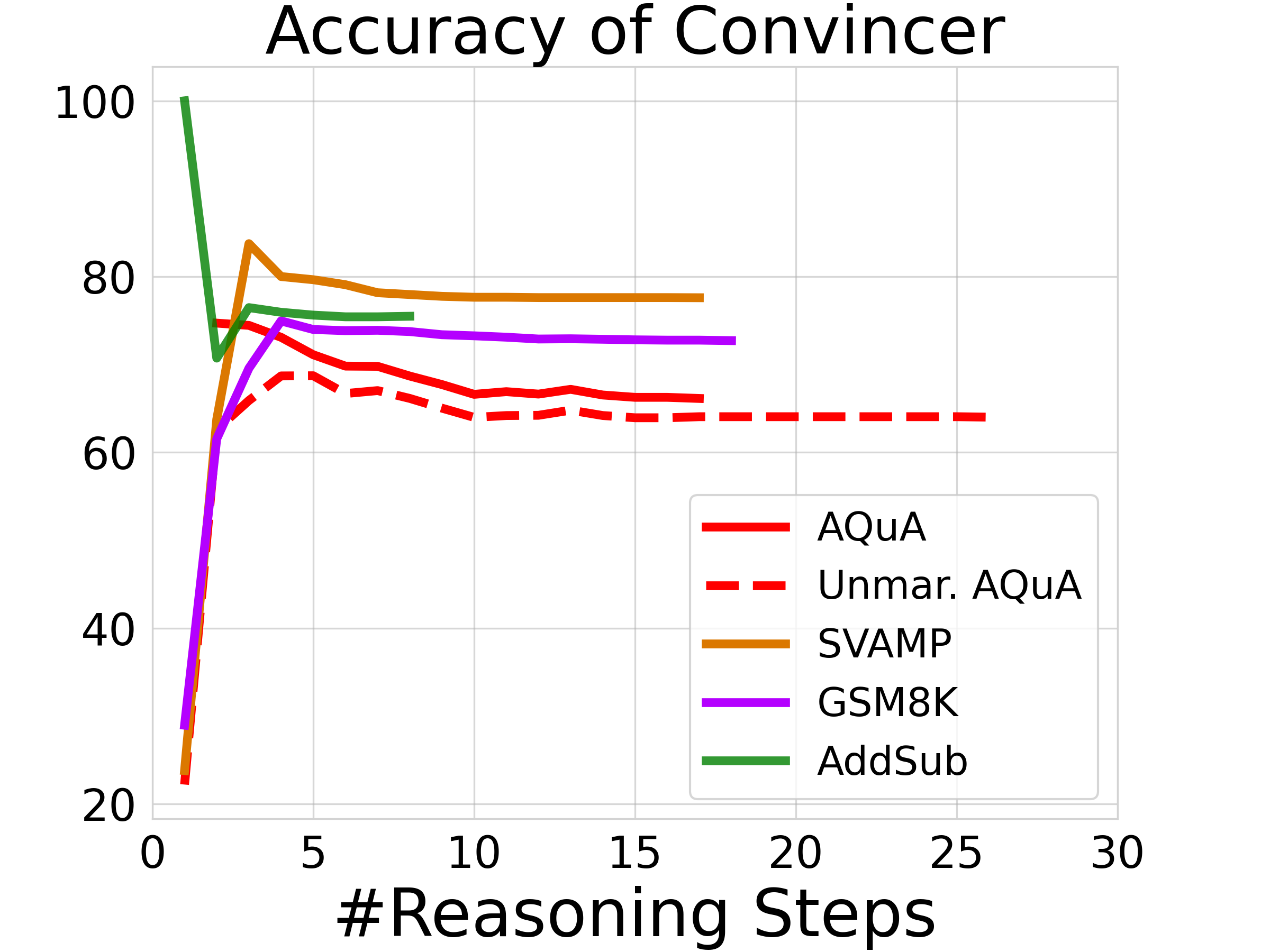}
\end{subfigure}
\caption{\textbf{Left:} performance with more iterations; \textbf{Middle:} transferring arithmetic exemplar to commonsense problems on different modules. \textbf{Right:} average accuracy of \textit{Convincer} on 5 iterations; accuracy of unmarginalized \textit{Convincer} on AQuA is also shown in the figure.}
\label{fig:more_iter}
\label{fig:transfer}
\label{fig:reasoning_steps}
\end{figure*}

\paragraph{Effectiveness of Self-Convince in Challenging Arithmetic Tasks}
When evaluating English arithmetic tasks, we can rank them by difficulty as follows: \textbf{``AQuA $>$ GSM8K $\approx$ SVAMP $>$ AddSub''}, which can also be simply measured by accurarcies of \textit{Convincer} shown in Figure~\ref{fig:reasoning_steps}. Table~\ref{table:main_result} reveals that \textit{Self-Convince} exhibits notable improvements across all types of inference. Particularly, it achieves the highest improvement on AQuA (+6.1\% on average), followed by GSM8K (+4.1\%), SVAMP (+4.0\%), and AddSub (+0.8\%)

\paragraph{Versatility and Potential of Self-Convince across Diverse Problem Domains}
While a slight degradation is observed in AddSub, which is consistent with the findings in PHP, our proposed method demonstrates significant performance enhancements across all other benchmarks when compared to the initial Manual CoT approach. Notably, our method achieves superior performance in arithmetic tasks, surpassing PHP even when the initial performance of Manual CoT is lower than that of PHP in SVAMP and AQuA. Moreover, the results obtained on the commonsense datasets provide compelling evidence of the potential applicability of our framework in broader problem domains.

\begin{figure*}[thb]
        \begin{subfigure}[t]{0.33\linewidth}
        \captionsetup{width=0.9\linewidth}
        	\centering
        	\includegraphics[width=\linewidth]{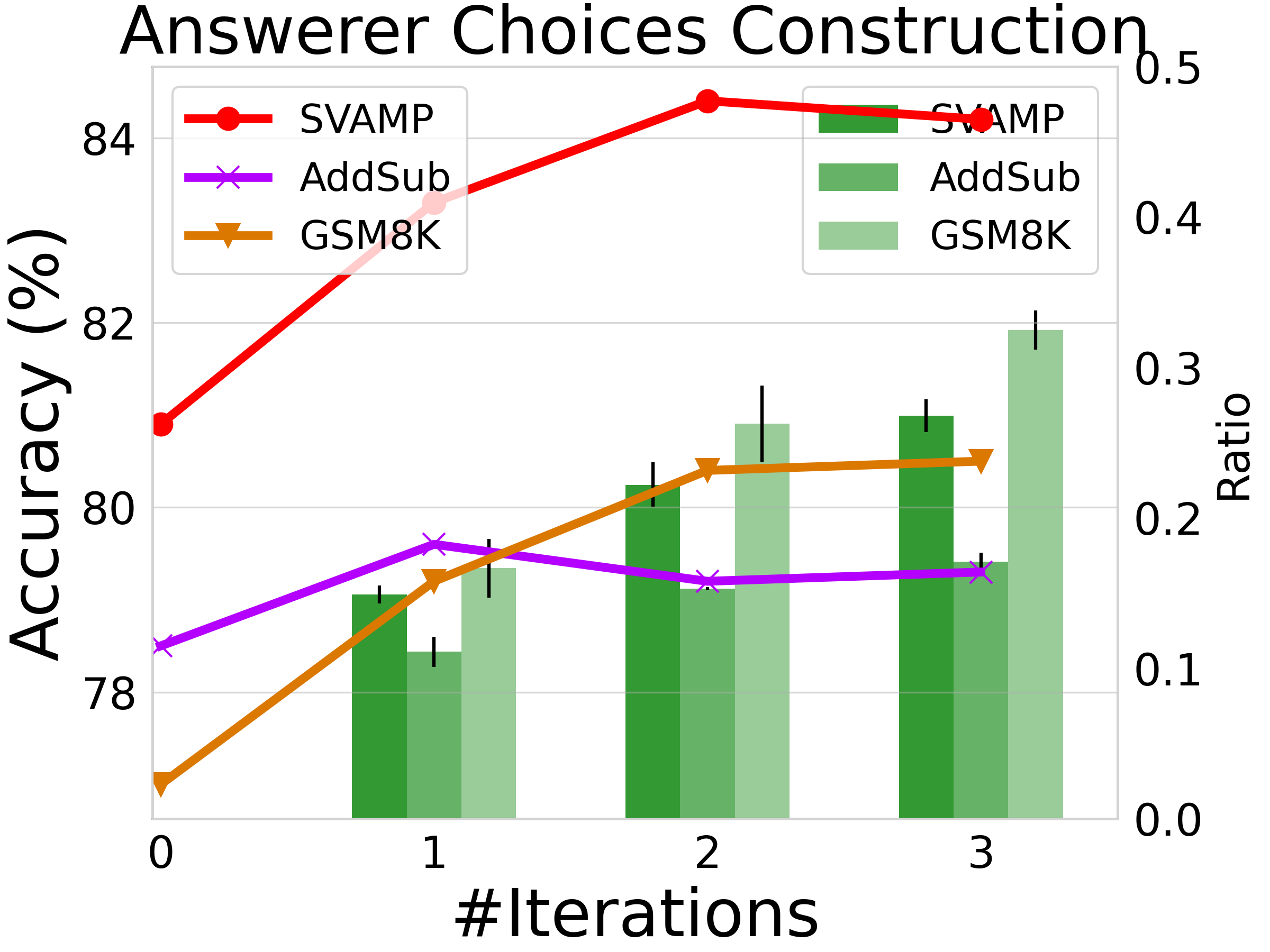}
        \end{subfigure}
        \begin{subfigure}[t]{0.33\linewidth}
        \captionsetup{width=0.9\linewidth}
        	\centering
        	\includegraphics[width=\linewidth]{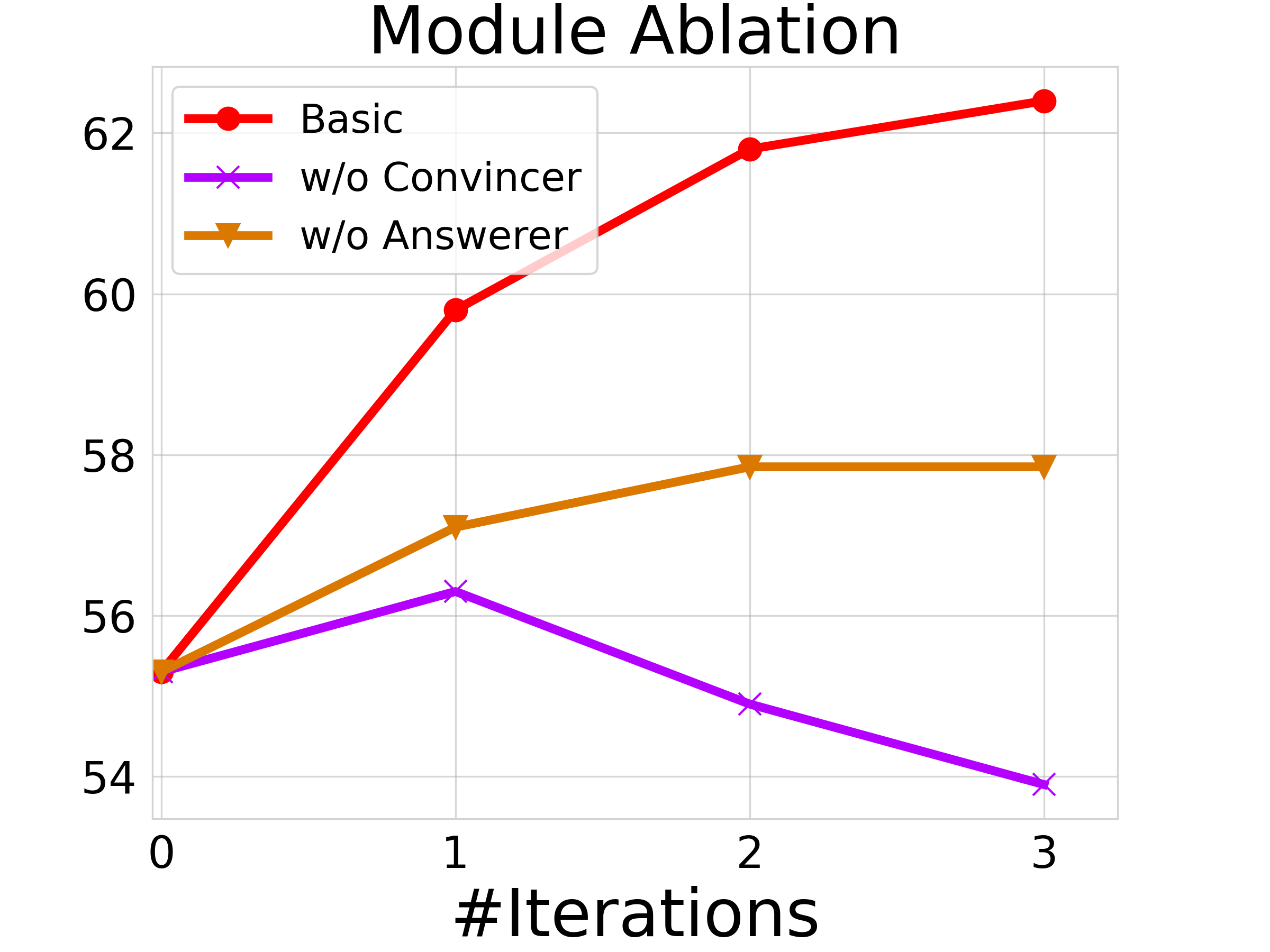}
        \end{subfigure}
    \begin{subfigure}[t]{0.33\linewidth}
    	\centering
    	\includegraphics[width=\linewidth]{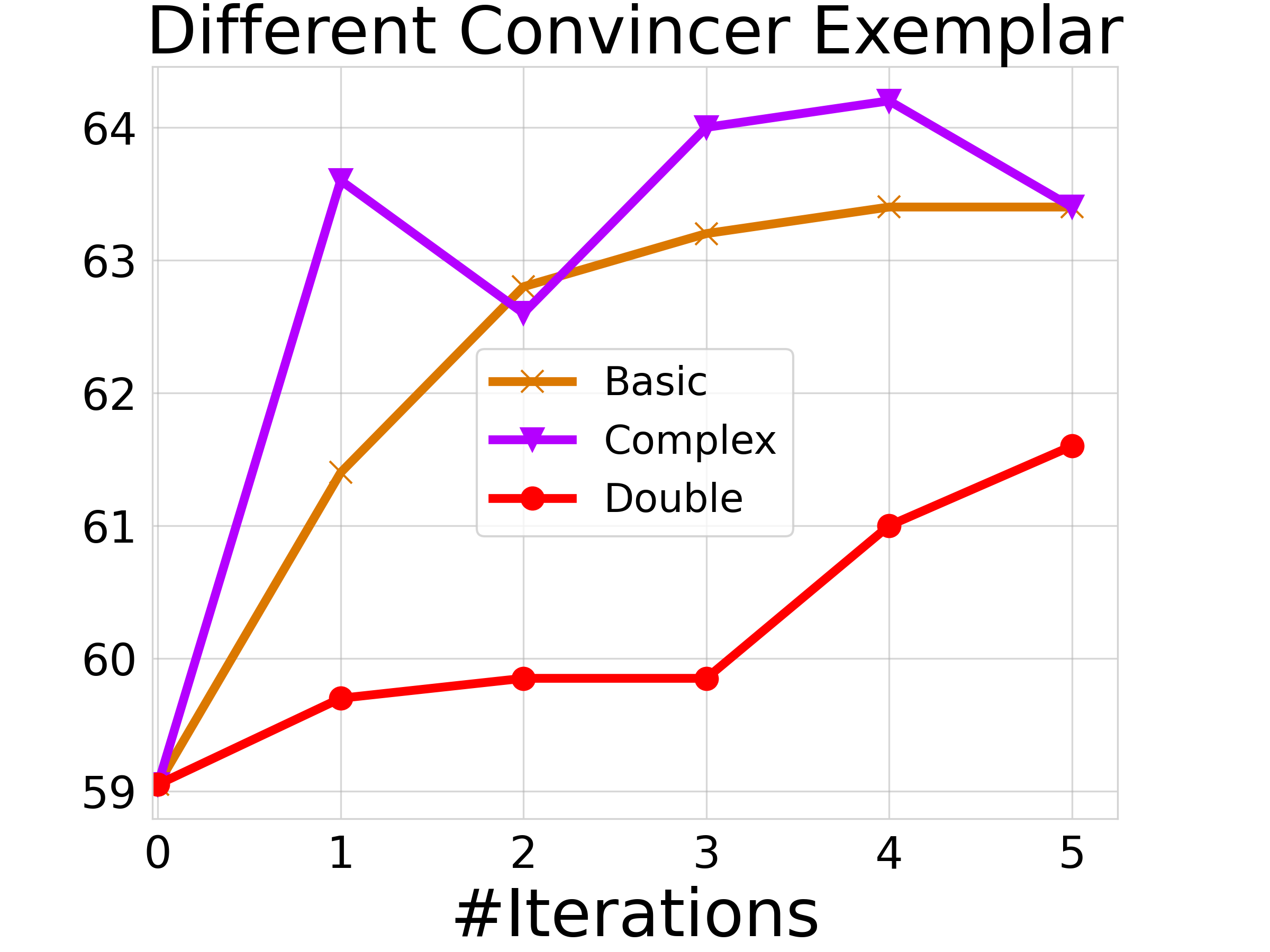}
    \end{subfigure}
    \caption{\textbf{Left:} accuracy on the benchmarks during answer choice construction, and the proportion of questions that have answer choices constructed. \textbf{Middle:} module ablation on \textit{Convincer} and \textit{Answerer}. \textbf{Right:} average accuracy of two runs with different exemplars of \textit{Convincer}; shadow areas depict the standard deviation.}
    \label{fig:answer_construction}
    \label{fig:module_ablation}
    \label{fig:convincer_exemplar}
\end{figure*}

\subsection{Ablation}
\label{sec:ablation}
In order to gain further insights into the effectiveness of \textit{Self-Convince}, we conducted a series of ablation experiments.
\paragraph{Answer Choice Construction}
Building upon the observed improvements in AQuA, we propose a variation of the \textit{Convincer} module called \textit{Wrong-only Convincer} to generate answer choices for other arithmetic datasets. The results presented in Table~\ref{table:main_result} demonstrate that the \textit{Wrong-only Convincer} alone is capable of enhancing the performance of LLMs, suggesting that self-hinted answer choices generally benefit the model's performance. Additionally, we provide results from different iterations of the \textit{Wrong-only Convincer} in Figure~\ref{fig:answer_construction}, along with the proportion of questions to which new options were added.
\paragraph{Convincer and Answerer}
Figure~\ref{fig:module_ablation} showcases the results on AQuA after removing either the \textit{Convincer} or the \textit{Answerer} module. The degradation in accuracy is evident when either module is omitted. Notably, the impact is more severe when the \textit{Convincer} module is removed compared to when the \textit{Answerer} module is removed. This indicates that in addition to providing supplementary information or reattempting incorrectly answered questions, the inclusion of extra information for incorrectly answered questions further enhances the performance of LLMs.
\paragraph{Different Convincer Exemplars}
To evaluate the impact of employing different manually-crafted exemplars within the \textit{Convincer} module, we conducted an experiment. Figure~\ref{fig:convincer_exemplar} showcases the results obtained by using three distinct settings: the \textbf{Basic} exemplar, an exemplar featuring more \textbf{Complex} examples, and a combination of two \textit{Convincers} utilizing different exemplars (\textbf{Double}). It is worth noting that the construction of exemplars holds significant importance in CoT prompting; however, the optimization of exemplar construction falls beyond the scope of this paper. As depicted in Figure~\ref{fig:convincer_exemplar}, the \textbf{Complex} exemplar exhibits a marginal enhancement in performance, whereas the adoption of the \textbf{Double} configuration results in a drastic decrease in accuracy. Furthermore, it is observed that all three exemplars exhibit a high standard deviation in the early iterations. With the exception of \textbf{Double}, the standard deviations progressively decrease with an increasing number of iterations.
\paragraph{Increased Number of Iterations}
Table~\ref{table:main_result} presents the results for up to 5 iterations. To further elucidate the effectiveness of longer iteration runs, we conducted additional experiments on the AQuA dataset, employing up to 10 iterations. These results are depicted in Figure~\ref{fig:more_iter}.
\paragraph{Transferring Arithmetic Exemplar to Commonsense Problems}
We explored the transferability of the \textit{Convincer} and \textit{Answerer} modules from arithmetic exemplars to commonsense tasks. The results depicted in Figure~\ref{fig:transfer} indicate that transferring either the \textit{Convincer} or the \textit{Answerer} module from arithmetic to commonsense tasks leads to performance improvements. However, when both modules are used with arithmetic exemplars, a significant drop in performance is observed.
\paragraph{Effectiveness of Convincer with Respect to Reasoning Steps}
Table~\ref{table:main_result} reveals that, for arithmetic benchmarks, the improvement relative to Manual CoT becomes more pronounced as the task difficulty increases. Therefore, we investigated how the accuracy of the \textit{Convincer} module varies with the number of reasoning steps. The number of correctly judged answers at each iteration is presented in Figure~\ref{fig:reasoning_steps}.

\section{Discussion and Limitations}
\label{sec:discussion}
In this section, we analyze the phenomena observed during our experiments and provide a discussion on these observations. Additionally, we identify certain limitations in our framework and present potential directions for future research.

\paragraph{The model has high accuracy on model-confident questions.}
In previous sections, we introduced the \textit{Wrong-only Convincer} to address the issues highlighted in Figure~\ref{fig:pilot_convincer}. However, we have observed that the model does not always generate new answers, and in some cases, it even asserts the correctness of an answer when explicitly informed that it is incorrect (\textit{``Correctness: Wrong''}). Consequently, it is not guaranteed that answer choices can be constructed for all questions. We refer to these instances as model-confident questions. Nonetheless, we acknowledge that the model's confidence in a wrong answer does not necessarily reflect its accuracy. 

\paragraph{Marginalization can slightly improve Convincer's accuracy.}
Despite the challenges associated with acquiring marginalization abilities, we have shown that the framework's design effectively improves the performance on AQuA. Moreover, we look into the accuracy of \textit{Convincer} with different designs. Results in Figure~\ref{fig:reasoning_steps} show that marginalization can obviously reduce reasoning steps in the ``Introspect'' phase, and slightly improve the accuracy of \textit{Convincer}.

\paragraph{The acquisition of the ability to marginalize and engage in step-wise reasoning poses significant challenges.}
The acquisition of the ability to marginalize subsequent reasoning steps and engage in step-wise reasoning presents significant challenges within our framework. In the design of the \textit{Convincer} module, our intention was for the model to rectify the initial mistake and subsequently marginalize the subsequent reasoning steps. However, upon careful examination of the model's behavior, we have observed instances where the model completes the entire reasoning process and provides an answer after rectifying the initial mistake.

Similarly, with the \textit{Answerer} module, our expectation was for it to engage in step-wise inference by generating additional reasoning steps based on the intermediate answer. However, in certain cases, the model either produces the complete reasoning process or solely outputs the type information before terminating. Consequently, under specific circumstances, our framework operates as an \textit{Answerer-first Inference} approach, where the input to $f_P$ is parsed as the portion following \textit{``A:''}, and the type information is parsed as the input when \textit{``A:''} is absent.

Specifically, approximately 30\% of the examples in the AQuA dataset fail in achieving marginalization, while approximately 35.4\% of the examples fail in generating a single reasoning step. 

\section{Conclusion}
In this paper, we propose a novel iterative framework named \textit{Self-Coninvce}. Our method show strong performance over 6 datasets, We also discover some promising features of our framework, e.g., using \textit{Wrong-only Convincer} to construct answer choices iteratively leads to better performance. For deeper exploration, we will leave it to future work.



\bibliography{emnlp}
\bibliographystyle{acl_natbib}


\clearpage

\appendix


\section{Appendix: CoT Prompts}\label{sec:appendix-full-prompts}

\subsection{Normal-CoT}

\begin{table*}[t]
    \centering
    \begin{tabular}{p{0.9\linewidth}}
        \toprule
        \textbf{Q:} There are 15 trees in the grove. Grove workers will plant trees in the grove today. After they are done, there will be 21 trees. How many trees did the grove workers plant today? \\
\textbf{A:}We start with 15 trees. Later we have 21 trees. The difference must be the number of trees they planted. So, they must have planted 21 - 15 = 6 trees. The answer is 6. \\
\textbf{Q:} If there are 3 cars in the parking lot and 2 more cars arrive, how many cars are in the parking lot? \\
\textbf{A:} There are 3 cars in the parking lot already. 2 more arrive. Now there are 3 + 2 = 5 cars. The answer is 5. \\
\textbf{Q:}  Leah had 32 chocolates and her sister had 42. If they ate 35, how many pieces do they have left in total? \\
\textbf{A:} Leah had 32 chocolates and Leah’s sister had 42. That means there were originally 32 + 42 = 74 chocolates. 35 have been eaten. So in total they still have 74 - 35 = 39 chocolates. The answer is 39. \\
\textbf{Q:} Jason had 20 lollipops. He gave Denny some lollipops. Now Jason has 12 lollipops. How many lollipops did Jason give to Denny? \\
\textbf{A:} Jason had 20 lollipops. Since he only has 12 now, he must have given the rest to Denny. The number of lollipops he has given to Denny must have been 20 - 12 = 8 lollipops. The answer is 8. \\
\textbf{Q:} Shawn has five toys. For Christmas, he got two toys each from his mom and dad. How many toys does he have now? \\
\textbf{A:} He has 5 toys. He got 2 from mom, so after that he has 5 + 2 = 7 toys. Then he got 2 more from dad, so in total he has 7 + 2 = 9 toys. The answer is 9.  \\
\textbf{Q:}  There were nine computers in the server room. Five more computers were installed each day, from monday to thursday. How many computers are now in the server room? \\
\textbf{A:} There are 4 days from monday to thursday. 5 computers were added each day. That means in total 4 * 5 = 20 computers were added. There were 9 computers in the beginning, so now there are 9 + 20 = 29 computers. The answer is 29. \\
\textbf{Q:} Michael had 58 golf balls. On tuesday, he lost 23 golf balls. On wednesday, he lost 2 more. How many golf balls did he have at the end of wednesday? \\
\textbf{A:} Michael initially had 58 balls. He lost 23 on Tuesday, so after that he has 58 - 23 = 35 balls. On Wednesday he lost 2 more so now he has 35 - 2 = 33 balls. The answer is 33. \\
\textbf{Q:} TOlivia has \$23. She bought five bagels for \$3 each. How much money does she have left? \\
\textbf{A:} She bought 5 bagels for \$3 each. This means she spent 5 * \$3 = \$15 on the bagels. She had \$23 in beginning, so now she has \$23 - \$15 = \$8. The answer is 8. \\
        \bottomrule
    \end{tabular}
    \caption{Normal-CoT-Arithmetic-except-AQuA}
    \label{tab:my_label1}
\end{table*}

\begin{table*}[t]
    \centering
    \begin{tabular}{p{0.9\linewidth}}
        \toprule
\textbf{Q:} John found that the average of 15 numbers is 40. If 10 is added to each number then the mean of the numbers is? Answer Choices: (a) 50 (b) 45 (c) 65 (d) 78 (e) 64 \\
\textbf{A:} If 10 is added to each number, then the mean of the numbers also increases by 10. So the new mean would be 50. The answer is (a). \\
\textbf{Q:} If a / b = 3/4 and 8a + 5b = 22,then find the value of a. Answer Choices: (a) 1/2 (b) 3/2 (c) 5/2 (d) 4/2 (e) 7/2 \\
\textbf{A:} If a / b = 3/4, then b = 4a / 3. So 8a + 5(4a / 3) = 22. This simplifies to 8a + 20a / 3 = 22, which means 44a / 3 = 22. So a is equal to 3/2. The answer is (b). \\
\textbf{Q:}  A person is traveling at 20 km/hr and reached his destiny in 2.5 hr then find the distance? Answer Choices: (a) 53 km (b) 55 km (c) 52 km (d) 60 km (e) 50 km \\
\textbf{A:} The distance that the person traveled would have been 20 km/hr * 2.5 hrs = 50 km. The answer is (e). \\
\vspace{0mm}
\textbf{Q:} Kirk sells cars. On two sales, Kirk has received commissions of \$850, and \$705, and he has one additional sale pending. If Kirk is to receive an average (arithmetic mean) commission of exactly \$820 on the three sales, then the third commission must be? Answer Choices: (a) \$895 (b) \$905 (c) \$910 (d) \$915 (e) \$945 \\
\textbf{A:} Let's think step by step. Joan found 72 seashells and 12 starfishes. She gave Alyssa some of her seashells. So we need to find out how many seashells she gave to Alyssa. Joan has 28 seashells. So she gave Alyssa 44 seashells. The answer is 44. \\
\textbf{Q:} Last year , 90171 people were born in a country , and 16320 people immigrated to it . How many new people began living in the country last year ? \\
\textbf{A:} Let x be the third value. So, 850 + 705 + x = 820 * 3. Solving for x: 1555 + x = 2460, x = 905. The answer is (b). \\
        \bottomrule
    \end{tabular}
    \caption{Normal-CoT-AQuA}
    \label{tab:my_label2}
\end{table*}

\begin{table*}[t]
    \centering
    \begin{tabular}{p{0.9\linewidth}}
        \toprule

\textbf{Q:} There are 15 trees in the grove. Grove workers will plant trees in the grove today. After they are done, there will be 21 trees. How many trees did the grove workers plant today? \\
\textbf{A:} We start with 15 trees. Later we have 21 trees. The difference must be the number of trees they planted. So, they must have planted 21 - 15 = 6 trees. The answer is 6. \\
\textbf{Q:} If there are 3 cars in the parking lot and 2 more cars arrive, how many cars are in the parking lot? \\
\textbf{A:} There are 3 cars in the parking lot already. 2 more arrive. Now there are 3 + 2 = 5 cars. The answer is 5. \\
\textbf{Q:}  Leah had 32 chocolates and her sister had 42. If they ate 35, how many pieces do they have left in total? \\
\textbf{A:} Leah had 32 chocolates and Leah’s sister had 42. That means there were originally 32 + 42 = 74 chocolates. 35 have been eaten. So in total they still have 74 - 35 = 39 chocolates. The answer is 39. \\
\textbf{Q:} Jason had 20 lollipops. He gave Denny some lollipops. Now Jason has 12 lollipops. How many lollipops did Jason give to Denny? \\
\textbf{A:} Jason had 20 lollipops. Since he only has 12 now, he must have given the rest to Denny. The number of lollipops he has given to Denny must have been 20 - 12 = 8 lollipops. The answer is 8. \\
\textbf{Q:} Shawn has five toys. For Christmas, he got two toys each from his mom and dad. How many toys does he have now? \\
\textbf{A:} He has 5 toys. He got 2 from mom, so after that he has 5 + 2 = 7 toys. Then he got 2 more from dad, so in total he has 7 + 2 = 9 toys. The answer is 9.  \\
\textbf{Q:} There were nine computers in the server room. Five more computers were installed each day, from monday to thursday. How many computers are now in the server room? \\
\textbf{A:} There are 4 days from monday to thursday. 5 computers were added each day. That means in total 4 * 5 = 20 computers were added. There were 9 computers in the beginning, so now there are 9 + 20 = 29 computers. The answer is 29. \\
\textbf{Q:} Michael had 58 golf balls. On tuesday, he lost 23 golf balls. On wednesday, he lost 2 more. How many golf balls did he have at the end of wednesday? \\
\textbf{A:} Michael initially had 58 balls. He lost 23 on Tuesday, so after that he has 58 - 23 = 35 balls. On Wednesday he lost 2 more so now he has 35 - 2 = 33 balls. The answer is 33. \\
\textbf{Q:} Olivia has \$23. She bought five bagels for \$3 each. How much money does she have left? \\
\textbf{A:}  She bought 5 bagels for \$3 each. This means she spent 5 * \$3 = \$15 on the bagels. She had \$23 in beginning, so now she has \$23 - \$15 = \$8. The answer is 8. \\
        \bottomrule
    \end{tabular}
    \caption{Normal-CoT-Commonsense-Date}
    \label{tab:my_label3}
\end{table*}

\begin{table*}[t]
    \centering
    \begin{tabular}{p{0.9\linewidth}}
        \toprule

\textbf{Q:} There are 15 trees in the grove. Grove workers will plant trees in the grove today. After they are done, there will be 21 trees. How many trees did the grove workers plant today? \\
\textbf{A:} We start with 15 trees. Later we have 21 trees. The difference must be the number of trees they planted. So, they must have planted 21 - 15 = 6 trees. The answer is 6. \\
\textbf{Q:} If there are 3 cars in the parking lot and 2 more cars arrive, how many cars are in the parking lot? \\
\textbf{A:} There are 3 cars in the parking lot already. 2 more arrive. Now there are 3 + 2 = 5 cars. The answer is 5. \\
\textbf{Q:}  Leah had 32 chocolates and her sister had 42. If they ate 35, how many pieces do they have left in total? \\
\textbf{A:} Leah had 32 chocolates and Leah’s sister had 42. That means there were originally 32 + 42 = 74 chocolates. 35 have been eaten. So in total they still have 74 - 35 = 39 chocolates. The answer is 39. \\
\textbf{Q:} Jason had 20 lollipops. He gave Denny some lollipops. Now Jason has 12 lollipops. How many lollipops did Jason give to Denny? \\
\textbf{A:} Jason had 20 lollipops. Since he only has 12 now, he must have given the rest to Denny. The number of lollipops he has given to Denny must have been 20 - 12 = 8 lollipops. The answer is 8. \\
\textbf{Q:} Shawn has five toys. For Christmas, he got two toys each from his mom and dad. How many toys does he have now? \\
\textbf{A:} He has 5 toys. He got 2 from mom, so after that he has 5 + 2 = 7 toys. Then he got 2 more from dad, so in total he has 7 + 2 = 9 toys. The answer is 9.  \\
\textbf{Q:} There were nine computers in the server room. Five more computers were installed each day, from monday to thursday. How many computers are now in the server room? \\
\textbf{A:} There are 4 days from monday to thursday. 5 computers were added each day. That means in total 4 * 5 = 20 computers were added. There were 9 computers in the beginning, so now there are 9 + 20 = 29 computers. The answer is 29. \\
\textbf{Q:} Michael had 58 golf balls. On tuesday, he lost 23 golf balls. On wednesday, he lost 2 more. How many golf balls did he have at the end of wednesday? \\
\textbf{A:} Michael initially had 58 balls. He lost 23 on Tuesday, so after that he has 58 - 23 = 35 balls. On Wednesday he lost 2 more so now he has 35 - 2 = 33 balls. The answer is 33. \\
Michael initially had 58 balls. He lost 23 on Tuesday, so after that he has 58 - 23 = 35 balls. On Wednesday he lost 2 more so now he has 35 - 2 = 33 balls. The answer is 33.
\textbf{Q:} Olivia has \$23. She bought five bagels for \$3 each. How much money does she have left? \\
\textbf{A:}  She bought 5 bagels for \$3 each. This means she spent 5 * \$3 = \$15 on the bagels. She had \$23 in beginning, so now she has \$23 - \$15 = \$8. The answer is 8. \\
        \bottomrule
    \end{tabular}
    \caption{Normal-CoT-Commonsense-CSQA}
    \label{tab:my_label4}
\end{table*}

\clearpage

\subsection{Convincer}

\begin{table*}[t]
    \centering
    \begin{tabular}{p{0.9\linewidth}}
        \toprule
\textbf{Q:} John found that the average of 15 numbers is 40. If 10 is added to each number then the mean of the numbers is? Answer Choices: (a) 50 (b) 45 (c) 65 (d) 78 (e) 64 \\
\textbf{A:} If 10 is added to each number, then the mean of the numbers also increases by 10. 40 + 10 = 45. So the new mean would be 45. The answer is (b). \\
\textbf{Correctness:} Wrong. \\
\textbf{Analysis:} he above answer is arithmetically incorrect, because 40 + 10 is not equal to 45, instead, 40 + 10 = 50. \\
\textbf{Final Answer:} If 10 is added to each number, then the mean of the numbers also increases by 10. 40 + 10 = 50. \\
\textbf{Q:} How many keystrokes are needed to type the numbers from 1 to 500? Answer Choices: (a) 1156 (b) 1392 (c) 1412 (d) 500 (e) 1788 \\
\textbf{A:}1-9 = 9 keystrokes. 10-99 = 90 keystrokes. 100-499 = 400 keystrokes. 500 = 1 keystroke. Total = 9+90+400+1 = 500 keystrokes. The answer is (d). \\
\textbf{Correctness:} Wrong. \\
\textbf{Analysis:} The answer is conceptually incorrect, because the keystroke differs according to the number of digits. \\
\textbf{Final Answer:} There are 9 one-digit numbers from 1 to 9. There are 90 two-digit numbers from 10 to 99. There are 401 three-digit numbers from 100 to 500. \\
\textbf{Q:} A person is traveling at 20 km/hr and reached his destiny in 2.5 hr then find the distance? Answer Choices: (a) 53 km (b) 55 km (c) 52 km (d) 60 km (e) 50 km \\
\textbf{A:} The distance that the person traveled would have been 24 km/hr * 2.5 hrs = 60 km. The answer is (d). \\
\textbf{Correctness:} Wrong. \\
\textbf{Analysis:} The answer is factually incorrect, because according to the question, the speed of the person is not 24 km/hr, instead, it should be 20 km/hr. \\
\textbf{Final Answer:} The distance that the person traveled would have been 20 km/hr * 2.5 hrs = 50 km. \\
\textbf{Q:} If a / b = 3/4 and 8a + 5b = 22,then find the value of a. Answer Choices: (a) 1/2 (b) 3/2 (c) 5/2 (d) 4/2 (e) 7/2 \\
\textbf{A:} If a / b = 3/4, then b = 4a / 3. So 8a + 5(4a / 3) = 22. This simplifies to 8a + 20a / 3 = 22, which means 44a / 3 = 22. So a is equal to 3/2. The answer is (b). \\
\textbf{Correctness:} Correct. \\
\textbf{Analysis:} The answer is correct. \\
\textbf{Final Answer:} If a / b = 3/4, then b = 4a / 3. So 8a + 5(4a / 3) = 22. This simplifies to 8a + 20a / 3 = 22, which means 44a / 3 = 22. So a is equal to 3/2. \\
\textbf{Q:} If a / b = 3/4 and 8a + 5b = 22,then find the value of a. Answer Choices: (a) 1/2 (b) 3/2 (c) 5/2 (d) 4/2 (e) 7/2\\
\textbf{A:} If a / b = 3/4, then b = 4a / 3. So 8a + 5(4a / 3) = 22. This simplifies to 8a + 20a / 3 = 22, 8a + 20a / 3 = 22, 8a + 20a / 3 = 22, 8a + 20a / 3 = 22, 8a + 20a / 3 = 22, 8a + 20a / 3 = 22, 8a + 20a / 3 = 22, 8a + 20a / 3 = 22, 8a + 20a / 3 = 22, 8a + 20a / 3 = 22, 8a + 20a / 3 = 22, 8a + 20a / 3 = 22, 8a + 20a / 3 = 22, 8a + 20a / 3 = 22, 8a + 20a / 3 = 22, 8a + 20a / 3 = 22, 8a + 20a /3 = 22, 8a + 20a / 3 = 22, \\
\textbf{Correctness:} Wrong. \\
\textbf{Analysis:} The answer is making nonsense, and it should be truncated. \\
\textbf{Final Answer:} If a / b = 3/4, then b = 4a / 3. So 8a + 5(4a / 3) = 22. \\
        \bottomrule
    \end{tabular}
    \caption{Convincer-Normal-AQuA}
    \label{tab:my_label5}
\end{table*}

\begin{table*}[t]
    \centering
    \begin{tabular}{p{0.9\linewidth}}
        \toprule
\textbf{Q:} A rectangular piece of cloth 2 feet wide was cut lengthwise into two smaller rectangular pieces. The shorter piece was one-third of the length of the longer of the 2 new pieces and had an area of 12 square feet. What was the length Q in feet of the original piece of cloth before cutting? \\
\textbf{A:} Let the length of the original piece of cloth be L. Then the length of the longer piece would be 2L/3. The length of the shorter piece would be L/3. The area of the shorter piece would be (L/3)(2) = 2L/3. So 2L/3 = 12. L = 18. The answer is 18. \\
\textbf{Correctness:} Wrong. \\
\textbf{Analysis:} The above answer has a ratio error, context 'The shorter piece was one-third of the length of the longer of the 2 new pieces' means that the length of the shorter piece would be L/4 instead of L/3. \\
\textbf{Final Answer:} Let the length of the original piece of cloth be L. Then the length of the shorter piece would be L/4.The length of the longer piece would be 3L/4. \\
\textbf{Q:} The ratio of A'2019s and B'2019s salary is 9 : 4. If A'2019s salary is increased by 15\%, then his total salary becomes 5175. What is the salary of B? \\
\textbf{A:}Let A'2019s salary be 9x and B'2019s salary be 4x. So 9x + 4x = 5175. This simplifies to 13x = 5175. So x = 400. So A'2019s salary is 9x = 3600 and B'2019s salary is 4x = 1600. The answer is 4500. \\
\textbf{Correctness:} Wrong. \\
\textbf{Analysis:} The above answer has a ratio error, A'2019s salary is increased by 15\%.That means his salary becomes 9x * (100+15)/100.So 10.35x + 4x = 5175. \\
\textbf{Final Answer:} Let the salaries of A and B be 9x and 4x.
\\ 9x * 115/100 = 5175
\\ x = 500
\\ salary of B = 500 * 4 = 2000. \\
\textbf{Q:} A wholesaler allows a discount of 20 \% on the list price to the retailer. The retailer sells at 5\% below the list price. If the customer pays Rs.19 for an article what profit is made by the retailer on it? \\
\textbf{A:} The retailer sells at 5\% below the list price. So the retailer's price is 95\% of the list price. The list price is 100\% of the wholesaler's price. So the retailer's price is 95\% of 100\% of the wholesaler's price. So the retailer's price is 95\% of the wholesaler's price. The wholesaler's price is 80\% of the list price. So the retailer's price is 95\% of 80\% of the list price. So the retailer's price is 76\% of the list price. The customer pays Rs.19 for an article. So the retailer's price is Rs.19. So the list price is Rs.19 / 0.76 = Rs.25. The profit is Rs.25 - Rs.19 = Rs.6. The answer is Rs.5.5.  \\
\textbf{Correctness:} Wrong. \\
\textbf{Analysis:}The above answer has a ratio error, because the retailer's price is 95\% of 100\% of the list price, instead of 95\% of 80\% of the list price. \\
\textbf{Final Answer:} If wholesaler's list price=100\%, Retailer's CP=80\% \ SP=95\%
\\If SP of 95\% of the retailer costs Rs.19 to customer,so its CP of 80\% will cost 80*19/95=16 \\

    \end{tabular}
    \label{tab:my_label6}
\end{table*}

\begin{table*}[t]
    \centering
    \begin{tabular}{p{0.9\linewidth}}
\textbf{Q:} Under a buy one get one free offer, with the selling price of a t-shirt of 4200. The shopkeeper says he got 33.33\% profit. What is cost price? \\
\textbf{A:} The cost price of one t-shirt is half of the selling price. So the cost price of one t-shirt is 4200 / 2 = 2100. The profit is 33.33\% of the cost price. So the profit is 33.33\% of 2100 = 699.63. The cost price is 2100 + 699.63 = 2799.63. The answer is 1675.04.  \\
\textbf{Correctness:} Wrong. \\
\textbf{Analysis:} The answer has a relation error, because selling price contains profit and cost price.When the profit is 699.63 and the selling price is 2100, the cost price is 2100 - 699.63 = 1575.04. \\
\textbf{Final Answer:} sp=4200
\\profit=33.33\%=(100+33.33)=133.33\
\\cp=(100*4200)/133.33
\\=3150.07 (cp for 2 t-shirt)
\\so cp for 1t-shirt= 1575.04 \\
\textbf{Q:} Money is paid into an account in portions of \$500 and paid out in portions of \$300. If after a month the account grew by \$1,200, what is the least possible number of transactions that could have been made?
 \\
\textbf{A:} The account grew by \$1,200, so the total amount of money paid in must have been \$1,200 more than the total amount of money paid out. The total amount of money paid in must have been a multiple of \$500, and the total amount of money paid out must have been a multiple of \$300. The only way to get a difference of \$1,200 is to have paid in \$1,500 and paid out \$300. The answer is 3.  \\
\textbf{Correctness:} Wrong. \\
\textbf{Analysis:} The above answer doesn't have a ratio error. However, it is inconsistent with the question, because the question is asking for the total number of transactions. Paying in \$1,500 means 3 transactions. Paying out \$300 means one.The total number is 4. \\
\textbf{Final Answer:} The account grew by \$1,200, so the total amount of money paid in must have been \$1,200 more than the total amount of money paid out. The total amount of money paid in must have been a multiple of \$500, and the total amount of money paid out must have been a multiple of \$300. The only way to get a difference of \$1,200 is to have paid in \$1,500 and paid out \$300. Paying in \$1,500 means 3 transactions. Paying out \$300 means one. \\
\textbf{Q:} If a / b = 3/4 and 8a + 5b = 22,then find the value of a.
 \\
\textbf{A:} If a / b = 3/4, then b = 4a / 3. So 8a + 5(4a / 3) = 22. This simplifies to 8a + 20a / 3 = 22, which means 44a / 3 = 22. So a is equal to 3/2. The answer is 3/2. \\
\textbf{Correctness:} Correct. \\
\textbf{Analysis:} The answer is correct. \\
\textbf{Final Answer:} If a / b = 3/4, then b = 4a / 3. So 8a + 5(4a / 3) = 22. This simplifies to 8a + 20a / 3 = 22, which means 44a / 3 = 22. So a is equal to 3/2. \\
        \bottomrule
    \end{tabular}
    \caption{Convincer-Complex-1-2}
    \label{tab:my_label7}
\end{table*}

\begin{table*}[t]
    \centering
    \begin{tabular}{p{0.9\linewidth}}
        \toprule
\textbf{Q:} If the ratio of the ages of Maya and Chhaya is 6:5 at present, and fifteen years from now, the ratio will get changed to 9:8, then find Maya's present age. Answer Choices: (a) 24 years old (b) 30 years old (c) 18 years old (d) 33 years old (e) Nones \\
\textbf{A:} Let Maya's present age be 6x years and Chhaya's present age be 5x years. So, after 15 years, Maya's age will be 6x + 15 years and Chhaya's age will be 5x + 15 years. So, the ratio of their ages will be (6x + 15) : (5x + 15) = 9:8. This means that 6x + 15 = 9x and 5x + 15 = 8x. So, x = 15. So, Maya's present age is 6x = 6 * 15 = 90 years. The answer is (d). \\
\textbf{Correctness:} Wrong. \\
\textbf{Analysis:} The above answer has a calculation error, because that the equation (6x + 15) : (5x + 15) = 9:8 means (6x + 15)/9=(5x + 15) / 8. \\
\textbf{Final Answer:} Let Maya's and Chhaya's present age is 6x and 5x respectively. And, (6x+15)/(5x+15) = 9/8; Or, 48x+120 = 45x = 135; Or, 3x = 15; Or, x = 5; Present age of Maya = 6x = 30. \\
\textbf{Q:} John found that the average of 15 numbers is 40. If 10 is added to each number then the mean of the numbers is? Answer Choices: (a) 50 (b) 45 (c) 65 (d) 78 (e) 64 \\
\textbf{A:} If 10 is added to each number, then the mean of the numbers also increases by 10. 40 + 10 = 45. So the new mean would be 45.The answer is (b). \\
\textbf{Correctness:} Wrong. \\
\textbf{Analysis:} The above answer has a computation error, because 40 + 10 is not equal to 45, instead, 40 + 10 = 50. \\
\textbf{Final Answer:} If 10 is added to each number, then the mean of the numbers also increases by 10. 40 + 10 = 50. So the new mean would be 50.\\
\textbf{Q:} A certain prosthodontist specializes in implanting gold and silver teeth in his patients' mouths. He charges \$650 for a gold tooth and \$325 for a silver tooth. If his total fees for implanting gold and silver teeth last week were \$15,925 in total, and he implanted five more gold teeth than silver teeth, how many teeth in total did he implant over the week? Answer Choices: (a)31 (b) 32 (c) 33(d) 34 (e) 35 \\
\textbf{A:}Let x be the number of gold teeth and y be the number of silver teeth. So x + y = total number of teeth. 650x + 325y = 15,925. x = y + 5. So x + y = y + 5 + y = 2y + 5. So 2y + 5 = total number of teeth. 650x + 325y = 15,925. 650(y + 5) + 325y = 15,925. 325y + 3250 + 325y = 15,925. 650y + 3250 = 15,925. 650y = 15,925 - 3250 = 12,675. y = 12,675 / 650 = 19.5. So x = y + 5 = 19.5 + 5 = 24.5. So total number of teeth = 2y + 5 = 2(19.5) + 5 = 39 + 5 = 44. The answer is (d).  \\
\textbf{Correctness:} Wrong. \\
\textbf{Analysis:}The above answer has a calculation error.650(y + 5) + 325y = 15,925.It means that 650y + 3250 + 325y = 15,925,instead of 325y + 3250 + 325y = 15,925. \\
\textbf{Final Answer:}  Let x be the number of gold teeth and y be the number of silver teeth. So x + y = total number of teeth. 650x + 325y = 15,925. x = y + 5. So x + y = y + 5 + y = 2y + 5. So 2y + 5 = total number of teeth. 650x + 325y = 15,925. 650y + 3250 + 325y = 15,925. 975y + 3250 = 15,925. 975y = 15,925 - 3250 = 12,675.\\

    \end{tabular}
    \label{tab:my_label8}
\end{table*}

\begin{table*}[t]
    \centering
    \begin{tabular}{p{0.9\linewidth}}
\textbf{Q:} Tanya's grandfather was 8 times older to her 16 years ago. He would be 3 times of her age 8 years from now. What was ratio of ages of Tanya and her grandfather 8 years ago. Answer Choices: (a) 11:50 (b) 11:51 (c) 11:53 (d) 11:54 (e) None of these \\
\textbf{A:} Let Tanya's age be x years. Then her grandfather's age would have been 8x years 16 years ago. So 8x - 16 = 3(x + 8). This simplifies to 5x = 64. So x = 12.8. So Tanya's age was 12.8 years and her grandfather's age was 8 * 12.8 = 102.4 years. The ratio of their ages was 12.8:102.4 = 11:51. The answer is (b).  \\
\textbf{Correctness:} Wrong. \\
\textbf{Analysis:} The answer has a calculation error, because the equation 8x - 16 = 3(x + 8) is wrong. Tanya's age is x years now, and was (x-16) 16 years ago. Grandfather's age was 8(x - 16) 16 years ago. So he's 8(x - 16) +16 now. Tanya's age will be (x + 8) years 8 years from now. Grandfather's age would be 3(x + 8) then. So he's 3(x + 8) - 8 now. The correct equation is 8(x - 16) + 16 = 3(x + 8) - 8. \\
\textbf{Final Answer:} Let Tanya's age be x years now.So she was (x-16) 16 years ago. Grandfather's age was 8(x - 16) 16 years ago.So he's 8(x - 16) +1 now.Tanya's age will be (x + 8) years 8 years from now.Grandfather's age would be 3(x + 8) then.So he's 3(x + 8) - 8 now. 8(x - 16) +16 = 3(x + 8) - 8. 8x - 112 = 3x + 16. \\
\textbf{Q:} Money is paid into an account in portions of \$500 and paid out in portions of \$300. If after a month the account grew by \$1,200, what is the least possible number of transactions that could have been made?  Answer Choices: (a) 3 (b) 4 (c) 5 (d) 6 (e) 7
 \\
\textbf{A:} The account grew by \$1,200, so the total amount of money paid in must have been \$1,200 more than the total amount of money paid out. The total amount of money paid in must have been a multiple of \$500, and the total amount of money paid out must have been a multiple of \$300. The only way to get a difference of \$1,200 is to have paid in \$1,500 and paid out \$300. The answer is (a).   \\
\textbf{Correctness:} Wrong. \\
\textbf{Analysis:}The above answer doesn't have a calculation error. However, it is inconsistent with the question, because the question is asking for the total number of transactions. Paying in \$1,500 means 3 transactions. Paying out \$300 means one. The total number is 4. \\
\textbf{Final Answer:} The account grew by \$1,200, so the total amount of money paid in must have been \$1,200 more than the total amount of money paid out. The total amount of money paid in must have been a multiple of \$500, and the total amount of money paid out must have been a multiple of \$300. The only way to get a difference of \$1,200 is to have paid in \$1,500 and paid out \$300. Paying in \$1,500 means 3 transactions. Paying out \$300 means one. The total number is 4.\\
\textbf{Q:} If a / b = 3/4 and 8a + 5b = 22,then find the value of a. Answer Choices: (a) 1/2 (b) 3/2 (c) 5/2 (d) 4/2 (e) 7/2
 \\
\textbf{A:} If a / b = 3/4, then b = 4a / 3. So 8a + 5(4a / 3) = 22. This simplifies to 8a + 20a / 3 = 22, which means 44a / 3 = 22. So a is equal to 3/2. The answer is (b). \\
\textbf{Correctness:} Correct. \\
\textbf{Analysis:} The answer is correct. \\
\textbf{Final Answer:} If a / b = 3/4, then b = 4a / 3. So 8a + 5(4a / 3) = 22. This simplifies to 8a + 20a / 3 = 22, which means 442a / 3 = 22. So a is equal to 3/2. The answer is (b). \\
        \bottomrule
    \end{tabular}
    \caption{Convincer-Complex-2-2}
    \label{tab:my_label9}
\end{table*}

\begin{table*}[t]
    \centering
    \begin{tabular}{p{0.9\linewidth}}
        \toprule
\textbf{Q:} 2015 is coming in 36 hours. What is the date one week from today in MM/DD/YYYY? \\
\textbf{A:} If 2015 is coming in 36 hours, then it is coming in 2 days. 2 days before 01/01/2015 is 12/29/2014, so today is 12/29/2014. \\
\textbf{Correctness:} Wrong. \\
\textbf{Analysis:} he above answer is incorrect, because there are 31 days in December, thus 2 days before 01/01/2015 is 12/30/2014, so today is 12/30/2014  \\
\textbf{Final Answer:} If 2015 is coming in 36 hours, then it is coming in 2 days. 2 days before 01/01/2015 is 12/30/2014, so today is 12/30/2014. \\
\textbf{Q:} The first day of 2019 is a Tuesday, and today is the first Monday of 2019. What is the date today in MM/DD/YYYY? \\
\textbf{A:}If the first day of 2019 was Tuesday, then 01/01/2019 was a Tuesday. Today is the first monday, would be five days later. \\
\textbf{Correctness:} Wrong. \\
\textbf{Analysis:}The above answer is incorrect, because if the first day of 2019 was Tuesday, then the first Monday should be six days later, instead of five days. \\
\textbf{Final Answer:} If the first day of 2019 was Tuesday, then 01/01/2019 was a Tuesday. Today is the first monday, would be six days later. \\
\textbf{Q:} The concert was scheduled to be on 06/01/1943, but was delayed by one day to today. What is the date 10 days ago in MM/DD/YYYY? \\
\textbf{A:} One day after 06/01/1943 is 06/02/1943, so today is 06/02/1943. 10 days before today is 05/23/1943. So the answer is 05/23/1943.  \\
\textbf{Correctness:} Correct. \\
\textbf{Analysis:}The above answer is correct. \\
\textbf{Final Answer:} One day after 06/01/1943 is 06/02/1943, so today is 06/02/1943. 10 days before today is 05/23/1943. So the answer is 05/23/1943. \\
\textbf{Q:} It is 4/19/1969 today. What is the date 24 hours later in MM/DD/YYYY? \\
\textbf{A:} Today is 04/19/1969. 24 hours later is two days after today, which would be 04/21/1969. So the answer is 04/21/1969.  \\
\textbf{Correctness:} Wrong. \\
\textbf{Analysis:} The above answer is incorrect. 24 hours later is one day after today, which would be 04/20/1969, instead of 04/21/1969. \\
\textbf{Final Answer:} Today is 04/19/1969. 24 hours later is one day after today, which would be 04/20/1969. \\
\textbf{Q:} Jane thought today is 3/11/2002, but today is in fact Mar 12, which is 1 day later. What is the date 24 hours later in MM/DD/YYYY? \\
\textbf{A:} Today is 03/12/2002. So the date 24 hours later will be 03/13/2002. So the answer is 03/13/2002.  \\
\textbf{Correctness:} Correct. \\
\textbf{Analysis:} The above answer is correct. \\
\textbf{Final Answer:} Today is 03/12/2002. So the date 24 hours later will be 03/13/2002. So the answer is 03/13/2002. \\
\textbf{Q:} Jane was born on the last day of Feburary in 2001. Today is her 16-year-old birthday. What is the date yesterday in MM/DD/YYYY?\\
\textbf{A:} The last day of February is the 28th, so Jane was born on 02/28/2001. Today is her 16-year old birthday, so today is 02/28/2017. So yesterday was 02/28/2017. So the answer is 02/28/2017. \\
\textbf{Correctness:} Wrong. \\
\textbf{Analysis:} The above answer is incorrect, because today is 02/28/2017, then yesterday was 02/27/2017, instead of 02/28/2017. \\
\textbf{Final Answer:} The last day of February is the 28th, so Jane was born on 02/28/2001. Today is her 16-year old birthday, so today is 02/28/2017. So yesterday was 02/27/2017. So the answer is 02/27/2017. \\
        \bottomrule
    \end{tabular}
    \caption{Convincer-Date}
    \label{tab:my_label10}
\end{table*}

\begin{table*}[t]
    \centering
    \begin{tabular}{p{0.9\linewidth}}
        \toprule
 \textbf{Q:} What do people use to absorb extra ink from a fountain pen? Answer Choices: (a) shirt pocket (b) calligrapher’s hand (c) inkwell (d) desk drawer (e) blotter \\
\textbf{A:} The answer must be an item that can absorb ink. Of the above choices, only desk drawers are used to absorb ink. So the answer is (e). \\
\textbf{Correctness:} Wrong. \\
\textbf{Analysis:} The above answer is incorrect. Of the above choices, blotters are used to absorb ink, instead of desk drawers. \\
\textbf{Final Answer:} The answer must be an item that can absorb ink. Of the above choices, only blotters are used to absorb ink. \\
\textbf{Q:} What home entertainment equipment requires cable? Answer Choices: (a) radio shack (b) substation (c) television (d) cabinet \\
\textbf{A:}The answer must require cable. Of the above choices, only substation requires cable. So the answer is (c). \\
\textbf{Correctness:} Wrong. \\
\textbf{Analysis:} The above answer is incorrect. Of the above choices, only television requires cable. \\
\textbf{Final Answer:} The answer must require cable. Of the above choices, only television requires cable.\\
\textbf{Q:} The fox walked from the city into the forest, what was it looking for? Answer Choices: (a) pretty flowers (b) hen house (c) natural habitat (d) storybook \\
\textbf{A:} The answer must be something in the forest. Of the above choices, only natural habitat is in the forest. So the answer is (b).  \\
\textbf{Correctness:} Correct. \\
\textbf{Analysis:}The above answer is correct. \\
\textbf{Final Answer:} The answer must be something in the forest. Of the above choices, only natural habitat is in the forest. So the answer is (b). \\
\textbf{Q:} Sammy wanted to go to where the people were. Where might he go? Answer Choices: (a) populated areas (b) race track (c) desert (d) apartment (e) roadblock \\
\textbf{A:} The answer must be a place with few people. Of the above choices, only desert has few people. So the answer is (c).  \\
\textbf{Correctness:} Wrong. \\
\textbf{Analysis:} The above answer is incorrect. Sammy wanted to go to where the people were, which means she wanted to go to a place with a lot of people. \\
\textbf{Final Answer:}The answer must be a place with a lot of people.\\

    \end{tabular}
\end{table*}

\begin{table*}[t]
    \centering
    \begin{tabular}{p{0.9\linewidth}}
\textbf{Q:} Where do you put your grapes just before checking out? Answer Choices: (a) mouth (b) grocery cart (c) super market (d) fruit basket (e) fruit market \\
\textbf{A:} The answer should be the place where grocery items are placed before checking out. Of the above choices, fruit basket can hold grocery items. So the answer is (b).  \\
\textbf{Correctness:} Wrong. \\
\textbf{Analysis:} The above answer is incorrect, because usually, grocery cart makes sense for holding grocery items. \\
\textbf{Final Answer:} The answer should be the place where grocery items are placed before checking out. Of the above choices, grocery cart makes sense for holding grocery items. \\
\textbf{Q:}  Google Maps and other highway and street GPS services have replaced what? Answer Choices: (a) united states (b) mexico (c) countryside (d) atlas \\
\textbf{A:} The answer must be something that used to do what Google Maps and GPS services do, which is to do online shopping. The answer is not given in the above choices. \\
\textbf{Correctness:} Wrong. \\
\textbf{Analysis:} The above answer is incorrect, because what Google Maps and GPS services do is to give directions. \\
\textbf{Final Answer:} The answer must be something that used to do what Google Maps and GPS services do, which is to give directions. \\
\textbf{Q:}  Before getting a divorce, what did the wife feel who was doing all the work? Answer Choices: (a) harder (b) anguish (c) bitterness (d) tears (e) sadness \\
\textbf{A:}The answer should be the feeling of someone getting divorced who was doing all the work. Of the above choices, the closest feeling is sadness. So the answer is (c). \\
\textbf{Correctness:} Wrong. \\
\textbf{Analysis:} The above answer is incorrect, because the according to the context, the wife was doing all the work. So, the closest feeling should be bitterness, instead of sadness. \\
\textbf{Final Answer:} The answer should be the feeling of someone getting divorced who was doing all the work. Of the above choices, the closest feeling is bitterness. \\
        \bottomrule
    \end{tabular}
    \caption{Convincer-CSQA}
    \label{tab:my_label11}
\end{table*}
\clearpage

\subsection{Answerer}
\begin{table*}[t]
    \centering
    \begin{tabular}{p{0.9\linewidth}}
        \toprule
\textbf{Q:} John found that the average of 15 numbers is 40. If 10 is added to each number then the mean of the numbers is? Answer Choices: (a) 50 (b) 45 (c) 65 (d) 78 (e) 64 \\
\textbf{Intermediate Answer:} If 10 is added to each number, then the mean of the numbers also increases by 10. \\
\textbf{Type:} Arithmetic. \\
\textbf{A:} If 10 is added to each number, then the mean of the numbers also increases by 10. 40 + 10 = 50. \\
\textbf{Q:} If a / b = 3/4 and 8a + 5b = 22,then find the value of a. Answer Choices: (a) 1/2 (b) 3/2 (c) 5/2 (d) 4/2 (e) 7/2 \\
\textbf{Intermediate Answer:} If a / b = 3/4, then b = 4a / 3. So 8a + 5(4a / 3) = 22. \\
\textbf{Type:} Simplification. \\
\textbf{A:} If a / b = 3/4, then b = 4a / 3. So 8a + 5(4a / 3) = 22. This simplifies to 8a + 20a / 3 = 22, which means 44a / 3 = 22. \\
\textbf{Q:} A person is traveling at 20 km/hr and reached his destiny in 2.5 hr then find the distance? Answer Choices: (a) 53 km (b) 55 km (c) 52 km (d) 60 km (e) 50 km \\
\textbf{Intermediate Answer:} The distance that the person traveled would have been 20 km/hr * 2.5 hrs = 50 km. \\
\textbf{Type:} Choice Decision. \\
\textbf{A:} The distance that the person traveled would have been 20 km/hr * 2.5 hrs = 50 km. The answer is (e). \\
\textbf{Q:} Kirk sells cars. On two sales, Kirk has received commissions of \$850, and \$705, and he has one additional sale pending. If Kirk is to receive an average (arithmetic mean) commission of exactly \$820 on the three sales, then the third commission must be? Answer Choices: (a) \$895 (b) \$905 (c) \$910 (d) \$915 (e) \$945 \\
\textbf{Intermediate Answer:} Let x be the third value. So, 850 + 705 + x = 820 * 3. \\
\textbf{Type:} Solving Equations. \\
\textbf{A:}  Let x be the third value. So, 850 + 705 + x = 820 * 3. Solving for x: 1555 + x = 2460, x = 905.\\
        \bottomrule
    \end{tabular}
    \caption{Answerer-Normal}
    \label{tab:my_label12}
\end{table*}

\begin{table*}[t]
    \centering
    \begin{tabular}{p{0.9\linewidth}}
        \toprule

\textbf{Q:} 2015 is coming in 36 hours. What is the date one week from today in MM/DD/YYYY? \\
\textbf{Intermediate Answer:} If 2015 is coming in 36 hours, then it is coming in 2 days. 2 days before 01/01/2015 is 12/30/2014, so today is 12/30/2014. \\
\textbf{Type:} Subtract from a date. \\
\textbf{A:} If 2015 is coming in 36 hours, then it is coming in 2 days. 2 days before 01/01/2015 is 12/30/2014, so today is 12/30/2014. So one week from today will be 01/05/2015. \\
\textbf{Q:} The first day of 2019 is a Tuesday, and today is the first Monday of 2019. What is the date today in MM/DD/YYYY? \\
\textbf{Intermediate Answer:}If the first day of 2019 was Tuesday, then 01/01/2019 was a Tuesday. Today is the first monday, would be six days later. \\
\textbf{Type:} Add to a date. \\
\textbf{A:} If the first day of 2019 was Tuesday, then 01/01/2019 was a Tuesday. Today is the first monday, would be six days later. So today is 01/07/2019. \\
\textbf{Q:} Jane was born on the last day of Feburary in 2001. Today is her 16-year-old birthday. What is the date yesterday in MM/DD/YYYY? \\
\textbf{Intermediate Answer:} The last day of February is the 28th, so Jane was born on 02/28/2001. \\
\textbf{Type:} Get today's date. \\
\textbf{A:} The last day of February is the 28th, so Jane was born on 02/28/2001. Today is her 16-year old birthday, so today is 02/28/2017. \\
        \bottomrule
    \end{tabular}
    \caption{Answerer-Date}
    \label{tab:my_label13}
\end{table*}

\begin{table*}[t]
    \centering
    \begin{tabular}{p{0.9\linewidth}}
        \toprule
 \textbf{Q:} What do people use to absorb extra ink from a fountain pen? Answer Choices: (a) shirt pocket (b) calligrapher’s hand (c) inkwell (d) desk drawer (e) blotter \\
\textbf{Intermediate Answer:} The answer must be an item that can absorb ink. \\
\textbf{Type:} Stationery Item. \\
\textbf{A:} The answer must be an item that can absorb ink. Of the above choices, only blotters are used to absorb ink. \\
\textbf{Q:} What home entertainment equipment requires cable? Answer Choices: (a) radio shack (b) substation (c) television (d) cabinet \\
\textbf{Intermediate Answer:} The answer must require cable. \\
\textbf{Type:} Electrical Device. \\
\textbf{A:} The answer must require cable. Of the above choices, only television requires cable. \\
\textbf{Q:} The fox walked from the city into the forest, what was it looking for? Answer Choices: (a) pretty flowers (b) hen house (c) natural habitat (d) storybook \\
\textbf{Intermediate Answer:} The answer must be something in the forest. \\
\textbf{Type:} Animal Behaviour. \\
\textbf{A:} The answer must be something in the forest. Of the above choices, only natural habitat is in the forest. \\
\textbf{Q:}  Sammy wanted to go to where the people were. Where might he go? Answer Choices: (a) populated areas (b) race track (c) desert (d) apartment (e) roadblock\\
\textbf{Intermediate Answer:} The answer must be a place with a lot of people. \\
\textbf{Type:} Public Area. \\
\textbf{A:} The answer must be a place with a lot of people. Of the above choices, only populated areas have a lot of people.\\
        \bottomrule
    \end{tabular}
    \caption{Answerer-CSQA}
    \label{tab:my_label14}
\end{table*}

\end{document}